%% file: main.tex
\documentclass[10pt,journal,compsoc]{IEEEtran}

\input{utils/preamble}

\patchcmd{\IEEEbiographynophoto}{\vskip 4\baselineskip plus 1fil minus 0\baselineskip}{\vskip 1\baselineskip plus 1fil minus 0\baselineskip}{}{\errmessage{biography skip patch failed}}

\begin{document}

\title{\ourmodel{}: Rethinking Bilateral Reference \\for High-Resolution Dichotomous Image Segmentation}

\author{
    Peng Zheng,
    Deng-Ping Fan,
    Binglu Wang,
    Nicu Sebe\\
    {\small\texttt{zhengpeng0108@gmail.com}}\vspace{-10pt}
\IEEEcompsocitemizethanks{
\IEEEcompsocthanksitem \textit{Deng-Ping Fan (corresponding author)} is with the College of Computer Science, Nankai University, Tianjin 300350, China (e-mail: \mbox{dengpfan@gmail.com}).
Peng Zheng and Nicu Sebe are with the Department of Information Engineering and Computer Science, University of Trento, Trento 38123, Italy.
Binglu Wang is with the School of Astronautics, Northwestern Polytechnical University, Xi'an 710072, China.
}%
}

\markboth{IEEE TRANSACTIONS ON PATTERN ANALYSIS AND MACHINE INTELLIGENCE}%
{Zheng \MakeLowercase{\textit{et al.}}: Rethinking Bilateral Reference for High-Resolution Dichotomous Image Segmentation}

\IEEEtitleabstractindextext{
\begin{abstract}
   \justifying
 Dichotomous image segmentation requires pixel-accurate foreground--background separation while preserving fine structures such as thin filaments, sharp edges, and intricate textures. \ourmodelVOne{} addressed this challenge through bilateral references, combining image patches and gradient information, but its inward reference loses local spatial structure and lacks semantic guidance, while its outward reference relies on single-channel gradient attention. We recast this paradigm as controlled information recovery---locality determines what spatial information survives, semantic selectivity what information enters, and focused boundary supervision where refinement occurs---and propose \ourmodel{} to implement it: (1) locality-preserving and semantically guided detail injection through space-to-depth transformation, overlapping patches, and a Semantically Guided Gate; (2) a learned boundary stream with focused supervision that generates full-channel residual attention without requiring boundary annotations; and (3) a dual-branch decoder integrating depth-wise boundary convolutions with large-kernel attention. A modernized framework---LayerNorm, GELU, the Lion optimizer, and a DINOv3 backbone---cuts training from 600 to 50 epochs. Extensive experiments demonstrate state-of-the-art performance on DIS5K, camouflaged object detection, and high-resolution salient object detection across all 13 evaluation sets. Code is available at \url{https://github.com/ZhengPeng7/BiRefNet-V2}.

\end{abstract}
}

\maketitle
\IEEEdisplaynontitleabstractindextext
\IEEEpeerreviewmaketitle

\begin{figure}[t]
\centering
\includegraphics[width=\linewidth]{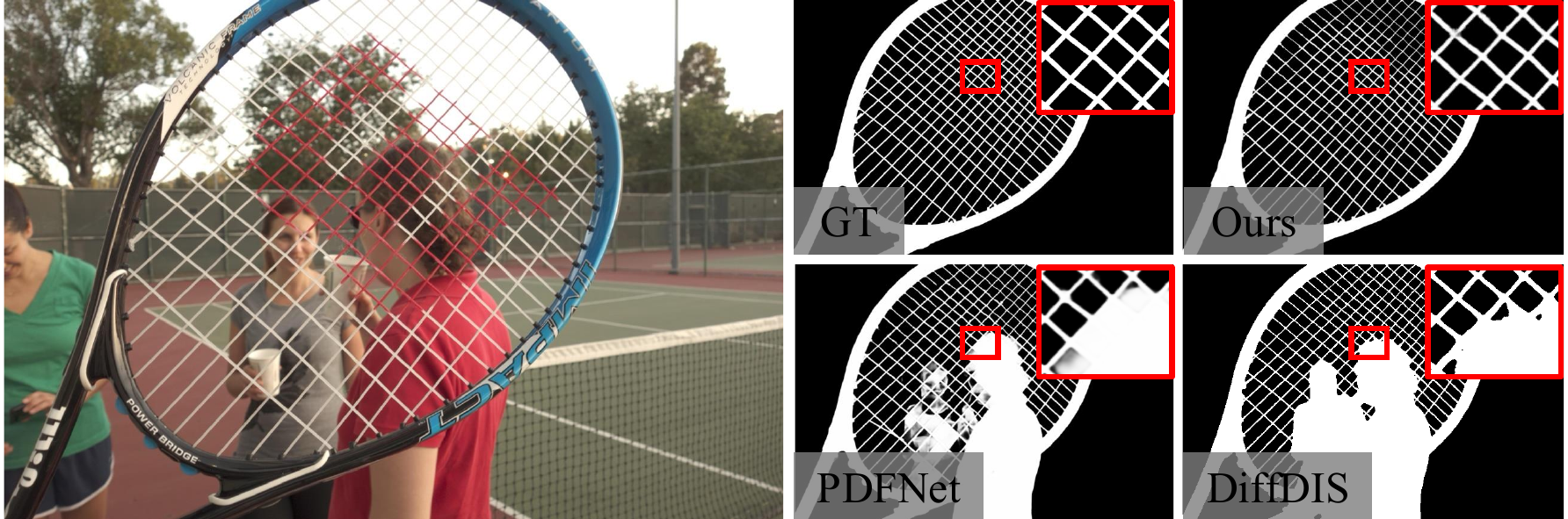}
\vspace{-20pt}
\caption{\textbf{Visual comparison on high-resolution dichotomous image segmentation.} \ourmodel{} versus PDFNet~\cite{PDFNet} and DiffDIS~\cite{DiffDIS} (mask binarized per its official protocol). Insets enlarge the red-boxed region. On a filament-level structure such as this racket's string bed, the competing methods blur the lattice into solid regions and absorb the people seen through the strings into the foreground, whereas \ourmodel{} reconstructs the topology nearly identically to the ground truth.}
\label{fig:teaser_qual}
\vspace{-11pt}
\end{figure}

\IEEEraisesectionheading{\section{Introduction}\label{sec:introduction}}
\IEEEPARstart{D}{ichotomous} image segmentation (DIS)~\cite{DIS5K} seeks masks that separate foreground from background with pixel-level accuracy in high-resolution natural images, where the objects of interest are defined by structures as fine as thin filaments, hollow lattices, and hair-level edges. Unlike salient object detection~\cite{SOC,DSS,PoolNet,EGNet,BASNet}, which tolerates coarse boundaries, DIS is judged by a complementary suite---the region-based maximum and weighted F-measure, the structure measure, and the human correction efforts $HCE_{\gamma}$---in which the maximum F-measure $F_{\beta}^{x}$ mainly reflects the large-area components of the mask while $HCE_{\gamma}$ counts the manual corrections that the fine structures demand. This evaluation regime sharpens a fundamental tension in encoder--decoder design: deep encoders must downsample aggressively to form the abstract representations needed for reliable foreground identification, yet the high-frequency content discarded during downsampling is exactly what $HCE_{\gamma}$ demands back at full resolution. Reconciling global semantic abstraction with lossless local detail is therefore the defining challenge of the field (\cref{fig:teaser_qual}).

\begin{figure*}[t]
\centering
\includegraphics[width=\textwidth]{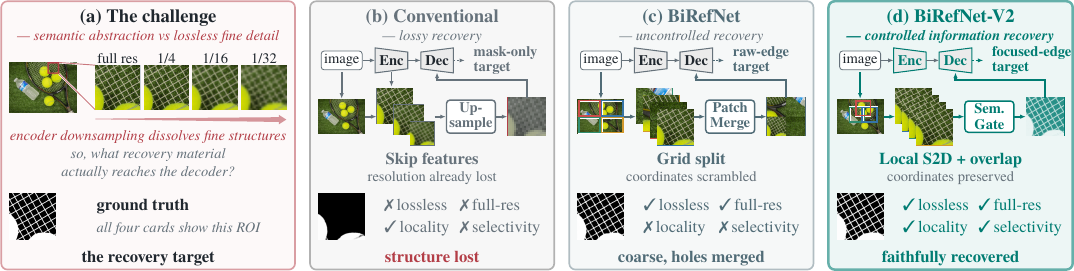}
\vspace{-20pt}
\caption{\textbf{The defining challenge of DIS and three architectural responses.} The mask tiles in (b)--(d) are real predictions, (b) by IS-Net~\cite{DIS5K}, on the red-boxed region of one DIS-TE4 image whose ground truth is given in (a); the remaining image content is rendered from the input pixels for illustration, except the teal map in (d), the RMS over channels of \ourmodel{}'s gated inward reference at full resolution, high (teal) between the strings and low (white) on them. Patch Merge: \ourmodelVOne{}'s channel-wise stacking of grid-split patches; S2D: space-to-depth; Sem.\ Gate: Semantically Guided Gate; focused-edge target: the prediction-gated image-edge target of \cref{sec:outward}.}
\label{fig:teaser_challenge}
\vspace{-10pt}
\end{figure*}

Existing approaches attack this tension from different perspectives, but each leaves a gap. Intermediate-supervision methods~\cite{DIS5K,UDUN} sharpen the encoder but never reinstate, during decoding, the detail lost to downsampling (\cref{fig:teaser_challenge}(b)); multi-scale aggregation schemes~\cite{MVANet,InSPyReNet} fuse features across resolutions, but the original image re-enters decoding at most as one ungated shallow branch added in the final upsampling steps, never as a reference available at every decoder stage. Recent diffusion-based methods~\cite{GenPercept,DiffDIS} import powerful generative priors at heavy inference cost, but still blur filament-level topology (DiffDIS in \cref{fig:teaser_qual}). \ourmodelVOne{}~\cite{BiRefNet} established the \emph{bilateral-reference} paradigm: an \emph{inward reference} that feeds original image content into every decoder stage to compensate for encoding loss, and an \emph{outward reference} that uses gradient cues to sharpen boundaries. The paradigm directly targets this tension.

Its \emph{original instantiation}, however, realizes only a fraction of the paradigm's potential. The inward reference splits the image into a grid of patches and stacks them along the channel dimension: every convolution response mixes content from grid cells that lie far apart in the image, and pixels adjacent across a patch border land at opposite spatial edges of the rearranged tensor, so the locality the reference aims to preserve is destroyed before any convolution sees it. The injection is moreover ungated, admitting raw texture everywhere with equal weight (\cref{fig:teaser_challenge}(c)). The outward reference, meanwhile, supervises its boundary head with an image-Laplacian target scaled by an unbounded intermediate logit rather than gated by a foreground probability, and applies the result as a bare single-channel rescaling that modulates all feature channels uniformly. These are not isolated defects to be patched, but symptoms that the paradigm has never been instantiated with its core requirements---locality, selectivity, and focused supervision---taken seriously.

In this paper we re-instantiate the bilateral-reference paradigm from the ground up: \ourmodel{} inherits the paradigm and the encoder--decoder skeleton from \ourmodelVOne{}, but its core components---both references and the decoder block that links them---are redesigned (\cref{tab:v1v2}). The redesign elevates those three requirements into a single principle---bilateral reference as \emph{controlled information recovery} (\cref{fig:teaser_challenge}(d)): detail recovery is a question of \emph{which information to inject and where}, controlled at both ends of the decoder---realized by the three components summarized below. Complementing these architectural changes, we switch the backbone from a fully fine-tuned Swin-L~\cite{Swin} to a self-supervised DINOv3~\cite{DINOv3} encoder kept \emph{mostly} frozen, with only its last transformer stage fine-tuned at a low learning rate (\cref{sec:training}), shrinking both the trainable stack and the training schedule while accuracy improves rather than degrades. The payoff is generality: trained separately per benchmark with no task-specific adaptation, the same architecture surpasses models purpose-built for camouflaged object detection and high-resolution salient object detection as well as for DIS---suggesting that these detail-demanding tasks pose the same information-recovery problem.

Our main contributions are summarized as follows:

\begin{itemize}

\item We formulate the conceptual advance over \ourmodelVOne{} as a general principle---bilateral reference as \emph{controlled information recovery}, with locality, selectivity, and focused supervision as its three requirements---and trace the failures of the original instantiation to these unmet requirements. This principle, not any single module, is our central contribution.

\item Implementing \emph{locality} and \emph{selectivity}, we re-instantiate the inward reference with locality-preserving detail injection---a space-to-depth transformation, overlapping patches with a lightweight compressor, and a Semantically Guided Gate---so that the decoder's own semantics decide where raw image detail enters.

\item Implementing \emph{focused supervision}, we upgrade the outward reference from single-channel image-gradient attention to full-channel residual attention driven by boundary features supervised with a prediction-gated image-edge target, making boundary refinement focused on the object being segmented---without any GT boundary labels.

\item Linking the two references structurally, we introduce a dual-branch decoder block that jointly models local boundaries (depth-wise convolutions) and global context (large-kernel attention).

\item Complementing the principle-driven redesign, we establish a modernized framework---LayerNorm, GELU, Lion, and a self-supervised DINOv3 encoder with a tailored fine-tuning scheme (\cref{sec:training})---and ablate it separately, so that its gains are never conflated with those of the paradigm.

\end{itemize}

Together, these improvements cut the trainable parameters to $\sim$40\% of \ourmodelVOne{}'s and accelerate training convergence by $12\times$ (\cref{sec:efficiency}).

\section{Related Work}
\label{sec:related}

\textbf{DIS.}
Qin~\etal~\cite{DIS5K} formalized DIS and introduced the DIS5K benchmark together with IS-Net, which mitigates the loss of fine structures during encoding via intermediate supervision. Later work attacked the same problem with frequency priors~\cite{FP-DIS}, a unite-divide-unite decomposition of trunk and boundary~\cite{UDUN}, image-pyramid processing at multiple resolutions~\cite{InSPyReNet}, and multi-view aggregation~\cite{MVANet}. Our prior work, \ourmodelVOne{}~\cite{BiRefNet}, introduced the \emph{bilateral-reference} paradigm.
More recently, generative priors have entered DIS. GenPercept~\cite{GenPercept} repurposes a pretrained diffusion model for dense perception tasks, including DIS, with single-step inference; DiffDIS~\cite{DiffDIS} adapts pretrained diffusion priors specifically to high-precision DIS, and is the strongest fully automatic diffusion-based method in our comparison; MaskFactory~\cite{MaskFactory} instead uses diffusion models offline to synthesize additional DIS training pairs. A separate line adds user prompts on top of generative modeling: LawDIS~\cite{LawDIS} formulates DIS as prompt-guided mask generation inside a latent diffusion model, and FlowDIS~\cite{FlowDIS} as text-guided flow matching that transports the image distribution to the mask distribution---a setting distinct from the fully automatic methods we compare against. SAMA~\cite{SAMA} and FCLM~\cite{FCLM} bring the same prompt-driven setting to discriminative models. Outside the generative family, PDFNet~\cite{PDFNet} pairs each image with a pseudo-depth map from a monocular depth estimator as a structural prior, and is the strongest of the published methods we compare against on $F_{\beta}^{x}$, the primary DIS metric.
Unlike the methods above, which strengthen the encoder, aggregate multiple scales, or rely on heavy generative priors, \ourmodel{} makes the original image itself re-enter every decoder stage---now under explicit semantic control---while keeping single-forward discriminative inference.

\textbf{Detail Preservation in High-Resolution Segmentation.}
Several families of methods compensate for the fine detail discarded by encoder downsampling: skip connections~\cite{UNet,FPN} forward encoder features to the decoder, but those features are already abstracted and lack pixel-level fidelity; multi-scale input schemes~\cite{InSPyReNet,ICNet} process the image at several resolutions in separate streams; and pyramid-grafting methods~\cite{PGNet} fuse features across lightweight and heavy backbones operating at different scales.
More directly, \ourmodelVOne{}~\cite{BiRefNet} injects original image patches into each decoder stage (the inward reference), and MVANet~\cite{MVANet} decomposes the input into multi-view patches within a single encoder--decoder stream. The injection of \ourmodelVOne{}, however, is indiscriminate and sacrifices spatial locality by stacking patches along the channel dimension; our redesigned inward reference addresses exactly these failures.

\textbf{Auxiliary Supervision for Feature Refinement.}
Beyond the primary segmentation loss, deep supervision~\cite{DIS5K,UDUN} applies the loss at multiple decoder stages, contour supervision~\cite{BGNet,EGNet} adds explicit boundary-prediction heads, and gradient-based supervision~\cite{DGNet} predicts image-derived edge maps to sharpen features near boundaries.
The outward reference of \ourmodelVOne{}~\cite{BiRefNet} follows the last route, but its Laplacian target responds to background texture as much as to object contours. \ourmodel{} instead gates the image-edge target with the decoder's evolving prediction and upgrades to full-channel residual attention (\cref{sec:outward}).

\section{Bilateral Reference Network V2}
\label{sec:method}

In this section, we present the architecture of \ourmodel{}. After the framework overview (\cref{sec:overview}), the structure mirrors the three requirements identified in \cref{sec:introduction}: \cref{sec:inward} restores \emph{locality} and adds \emph{selectivity} on the input side, \cref{sec:decoder_block} structurally links the two references, \cref{sec:outward} brings \emph{focused supervision} to the output side, and \cref{sec:training} details training and inference.

\subsection{Framework Overview}
\label{sec:overview}

\begin{figure*}[t]
\centering
\includegraphics[width=\linewidth]{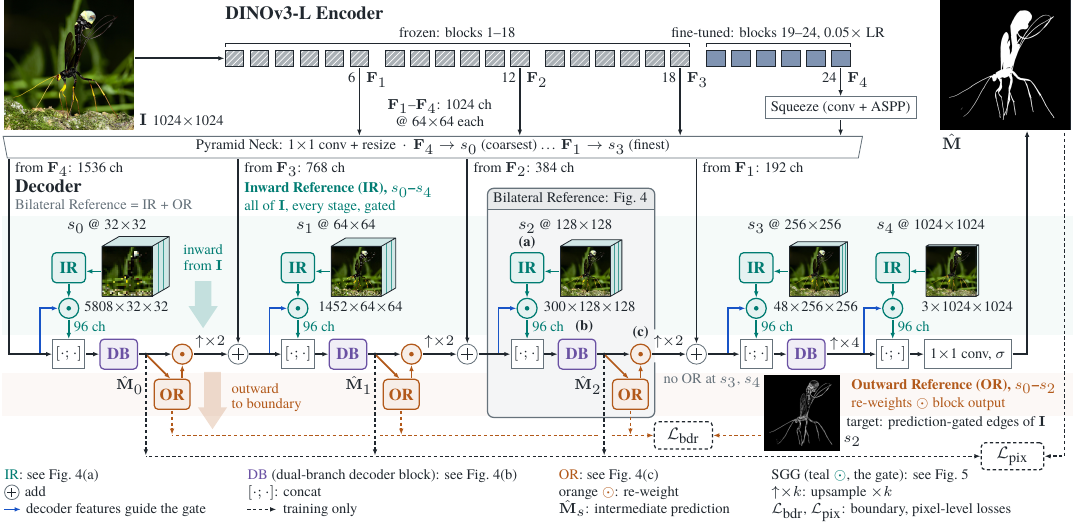}
\vspace{-20pt}
\caption{\looseness=-1\textbf{Overall architecture of \ourmodel{}.} The framed stage is detailed in \cref{fig:biref}, its gate in \cref{fig:sgg}; the edge target is square-root scaled for display.}
\label{fig:overall}
\vspace{-5pt}
\end{figure*}

\ourmodel{} follows an encoder--decoder architecture with bilateral reference. The pipeline is summarized below and shown in~\cref{fig:overall}.

\textbf{Encoder.}
We adopt a DINOv3-Large (DINOv3-L)~\cite{DINOv3} backbone encoder, a ViT-L/16 model with 24 transformer blocks. Given an input image $\mathbf{I}\in\mathbb{R}^{3\times H\times W}$, the backbone extracts four intermediate features $\{\mathbf{F}_i\}_{i=1}^{4}$ from transformer blocks $\{6,12,18,24\}$, ordered from shallowest to deepest. These four tapped blocks partition the otherwise uniform encoder into four consecutive block groups, which we call its \emph{stages}; the encoder is kept frozen except for the last of them---blocks $19$--$24$---which is fine-tuned as described in \cref{sec:training}. Since all four feature maps share the same channel dimension $C_\text{bb}=1024$ (a property of plain ViT architectures), we employ a \emph{pyramid neck} consisting of four $1\times1$ convolutions to project them into a hierarchical channel structure of $192$, $384$, $768$, and $1536$ channels for $\mathbf{F}_1$ to $\mathbf{F}_4$, and resize each to $\frac{H}{2^{i+1}} \times \frac{W}{2^{i+1}}$, restoring the multi-scale structure expected by FPN-style decoders~\cite{BiRefNet}.

\textbf{Squeeze module.}
Following \ourmodelVOne{}~\cite{BiRefNet}, we apply a squeeze module to the deepest feature $\mathbf{F}_4$ prior to the pyramid-neck projection. This module is a residual convolutional block with an Atrous Spatial Pyramid Pooling (ASPP)~\cite{DeepLabv3p} module between its two convolutions, which captures multi-scale context to enrich global semantics.

\textbf{Decoder.}
The decoder has four stages that progressively refine predictions from the coarsest to the finest resolution. At each stage $s\in\{0,1,2,3\}$ (written $s_0$--$s_3$, from deepest to shallowest, as in~\cref{fig:overall}), the processing follows three steps:
\begin{enumerate}
    \item \textbf{Inward reference (IR)}: The original image $\mathbf{I}$ is partitioned into patches at a resolution matching the current stage, processed by a lightweight convolutional block, and selectively injected into decoder features via a Semantically Guided Gate (SGG).
    \item \textbf{Dual-branch decoder block (DB)}: The concatenated features are transformed by a block that combines the modeling of local boundaries with the aggregation of global context.
    \item \textbf{Outward reference (OR)} (stages $s_0$--$s_2$): The decoder output is refined through full-channel residual attention, where the model learns to predict boundary maps supervised by prediction-gated image-edge targets during training.
\end{enumerate}

After stage~$s_3$, the decoder output is upsampled to the input resolution and concatenated with a final inward reference (the full-resolution injection site, denoted stage~$s_4$, where each patch is a single pixel) into $\mathbf{D}_\text{out}$ to produce the prediction $\hat{\mathbf{M}} = \sigma(\text{Conv}_{1\times1}(\mathbf{D}_\text{out})) \in [0,1]^{H\times W}$, where $\sigma$ denotes the sigmoid function.

\textbf{Multi-scale supervision.}
During training, intermediate predictions are generated at stages $s_0$--$s_2$ via $1\times1$ convolutions. Together with the final prediction, these form a set of multi-scale supervision signals, each bilinearly upsampled to the ground-truth resolution and supervised there. \cref{tab:v1v2} summarizes the component-level differences from \ourmodelVOne{}.

\begin{table}[t]
\centering
\caption{\textbf{Component-level changes from \ourmodelVOne{} to \ourmodel{}.} LKA: large-kernel attention; deform.: deformable convolution}
\label{tab:v1v2}
\vspace{-5pt}
\renewcommand{\arraystretch}{1.0}
\setlength{\tabcolsep}{2pt}
\scriptsize
\begin{tabular*}{\linewidth}{@{\extracolsep{\fill}} lll@{}}
\toprule
Component & \ourmodelVOne{} & \ourmodel{} \\
\midrule
Backbone            & Swin-L, fine-tuned      & DINOv3-L, last-stage tuned \\
\arrayrulecolor{black!25}\midrule
Norm / activation   & BatchNorm / ReLU        & LayerNorm / GELU \\
\midrule
Inward Reference    & grid-split, ungated     & S2D$+$overlap, SGG-gated \\
\midrule
Outward Reference   & image grad., 1-ch.      & pred.-gated edge, full-ch. \\
\midrule
Decoder block       & conv ($+$ deform.)      & dual-branch conv$+$LKA \\
\midrule
Optimizer / schedule & Adam, 600 epochs      & Lion, 50 epochs \\
\midrule
Trainable params    & 215M              & 84.9M \\
\arrayrulecolor{black}\bottomrule
\end{tabular*}
\vspace{-10pt}
\end{table}

\subsection{Inward Reference}
\label{sec:inward}

\begin{figure*}[t]
\centering
\includegraphics[width=\linewidth]{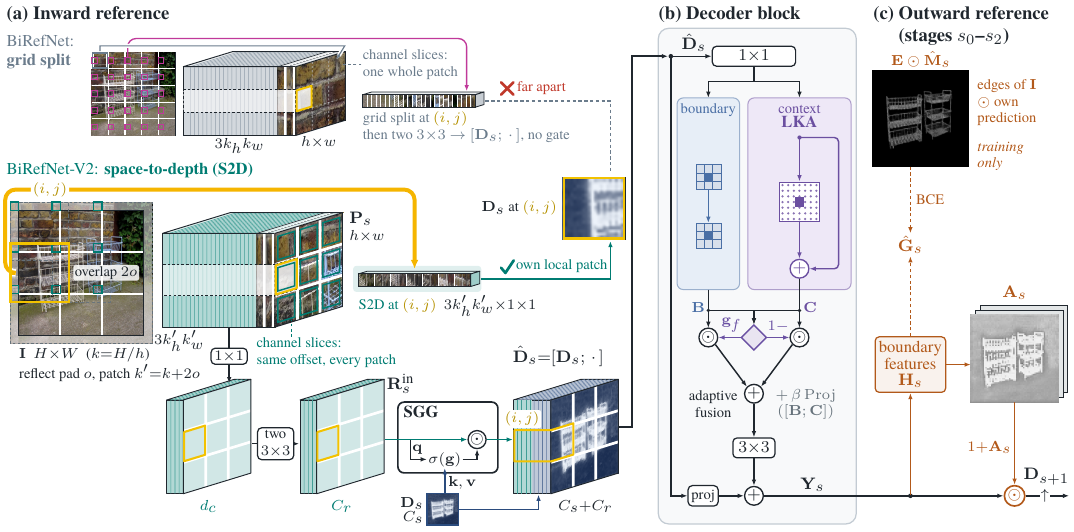}
\vspace{-20pt}
\caption{\textbf{Bilateral reference at one decoder stage.} Panel (a) is drawn on toy grids ($15{\times}15$ pixels, $3{\times}3$ positions), with an overlap border of $o{=}\lfloor\alpha k\rfloor$ pixels. Maps come from the trained model on one DIS-VD image: $\mathbf{D}_s$ is shown by its first principal component, $\mathbf{A}_s$ by its mean over channels, and $\mathbf{E}\odot\hat{\mathbf{M}}_s$ with a $0.5$ display gamma.}
\label{fig:biref}
\end{figure*}

Compared to \ourmodelVOne{}~\cite{BiRefNet}, our redesign addresses three limitations: broken spatial locality, patch boundary artifacts, and missing semantic guidance. \cref{fig:biref} illustrates one decoder stage with the redesigned module, the decoder block, and the outward reference.

\subsubsection{Locality-Preserving Patch Decomposition}

At each decoder stage $s$, we decompose the input image into patches whose spatial extent matches the downsampling grid cell of the corresponding feature map. For an input of size $H\times W$ and a feature map of size $h\times w$, each patch covers a $k_h\times k_w$ region with $k_h=H/h$ and $k_w=W/w$. We rearrange these patches into the channel dimension via a \emph{space-to-depth} (S2D) transformation:
\begin{equation}
    \mathbf{P}_s = \text{S2D}(\mathbf{I}) \in \mathbb{R}^{(3 \cdot k_h' \cdot k_w') \times h \times w},
\end{equation}
where $k_h'$ and $k_w'$ denote the effective patch size after accounting for overlap (detailed below). Unlike \ourmodelVOne{}'s grid split, our formulation assigns each spatial location the content of its own local patch: pixels that are neighbors in $\mathbf{I}$ remain neighbors within each patch, enabling subsequent convolutions to exploit local correlations effectively.
Each channel slice holds the same pixel offset of every patch, \ie, one phase of $\mathbf I$; the channel vector at $(i,j)$ holds that position's whole local patch, so $\mathbf I$ is recovered exactly.

\subsubsection{Overlapping Patches with Compression}

Non-overlapping patch boundaries introduce discontinuities at the edges of adjacent patches. To mitigate this, we extend each patch by a border of $\lfloor \alpha \cdot k \rfloor$ pixels on each side, taken from the neighboring patches (by reflection at the image border), where $\alpha=0.2$ is the overlap ratio. The patches are thus extracted with overlapping sliding windows of size $k'=k+2\lfloor\alpha \cdot k\rfloor$ for $k\in\{k_h,k_w\}$, yielding a higher channel count $3\cdot k_h' \cdot k_w'$; the border vanishes at the two finest stages, where $\lfloor\alpha \cdot k\rfloor=0$.
By contrast, \ourmodelVOne{}'s grid split is a bijective tensor reshape that cannot overlap patches at all, so $\alpha{>}0$ is reachable only by space-to-depth.

Since the expanded channel dimension can be large (\eg, $3\times44\times44=5808$ at the deepest stage), we apply a lightweight $1\times1$ convolution with LayerNorm and GELU that compresses the patches to a fixed dimension $d_c=64$ whenever their channel count exceeds it (in our configuration, the three coarsest stages). The compressed $\mathbf{P}_s$ is then passed through a two-layer convolutional block (\ie, two $3\times3$ convolutions with LayerNorm and GELU) to produce the inward-reference features $\mathbf{R}_s^{\text{in}} \in \mathbb{R}^{C_r \times h \times w}$, where $C_r=96$ for all stages.

\subsubsection{Semantically Guided Gate}

Ungated concatenation of $\mathbf{R}_s^{\text{in}}$ with the decoder features injects raw image content everywhere, regardless of what the decoder already knows about each position. We introduce a \emph{Semantically Guided Gate} that conditions the injection on the decoder's semantic context, so that the decoder itself decides where raw detail enters. In the trained model this decision is made at the full-resolution stage, the coarser gates varying little or not at all between foreground and background; at that stage the gate averages 0.12 inside the foreground, 0.40 along its boundary, and 0.63 in the background over the 470 DIS-VD images, with the background above the foreground in all but one (see the~\mbox{supplementary}~material).

Given the decoder features $\mathbf{D}_s$ and the inward-reference features $\mathbf{R}_s^{\text{in}}$, the SGG computes:
\begin{equation}
    \text{SGG}(\mathbf{D}_s, \mathbf{R}_s^{\text{in}}) = \mathbf{R}_s^{\text{in}} \odot \sigma(\mathbf{g}),
\end{equation}
where $\odot$ denotes element-wise multiplication and $\mathbf{g}$ is the gating signal computed below.

\begin{figure}[t]
\centering
\includegraphics[width=\linewidth]{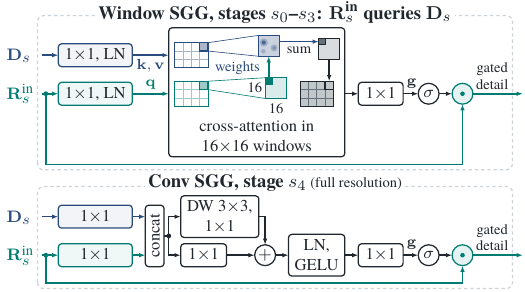}
\vspace{-20pt}
\caption{\textbf{The Semantically Guided Gate (SGG).} The gate $\sigma(\mathbf{g})$ has one value per channel and pixel of $\mathbf{R}_s^{\text{in}}$.}
\label{fig:sgg}
\end{figure}

Specifically, after $1\times1$ projections with LayerNorm to a shared attention dimension, $\mathbf{R}_s^{\text{in}}$ serves as queries $\mathbf{q}$ and $\mathbf{D}_s$ as keys and values $\mathbf{k},\mathbf{v}$ in a window-based cross-attention (\emph{Window SGG}, \cref{fig:sgg}; no shifted windows or positional encoding) with a fixed window size $w_s{=}16$, which yields $O(N \cdot w_s^2)$ complexity for $N$ spatial positions. The gating signal is then computed as:
\begin{equation}
    \mathbf{g} = \text{GateProj}(\text{OutProj}(\text{WindowAttn}(\mathbf{q}, \mathbf{k}, \mathbf{v}))),
\end{equation}
where WindowAttn is scaled dot-product attention within non-overlapping $w_s\times w_s$ windows; OutProj and GateProj are $1\times1$ convolutions (one box in \cref{fig:sgg}). At the final stage (input resolution), where the feature map is at $H\times W$, we use a lightweight convolutional gate (\emph{Conv SGG})---window attention at full resolution costs about a third more peak training memory per sample---that aligns both inputs via $1\times1$ convolutions and fuses them through parallel $1\times1$ and depth-wise separable $3\times3$ branches.

The gate is initialized close to open, so that the inward reference passes through nearly unmodified at the start of training. The gated inward reference is concatenated with the decoder features $\mathbf{D}_s$ along the channel dimension:
\begin{equation}
    \hat{\mathbf{D}}_s = [\mathbf{D}_s; \text{SGG}(\mathbf{D}_s, \mathbf{R}_s^{\text{in}})] \in \mathbb{R}^{(C_s + C_r) \times h \times w},
\end{equation}
where $[\cdot;\cdot]$ denotes channel-wise concatenation and $C_s$ is the decoder channel dimension at stage $s$ (width of $\mathbf{F}_{4-s}$ after the pyramid neck for $s_0$--$s_3$, from $1536$ to $192$, and $96$ at $s_4$).

\subsection{Dual-Branch Decoder Block}
\label{sec:decoder_block}

DIS requires simultaneously capturing fine boundary details and maintaining global structural coherence. A single processing pathway struggles to satisfy both requirements: small-kernel convolutions excel at local edge detection but lack context, while softmax self-attention captures long-range dependencies only at a quadratic cost that becomes prohibitive at high-resolution decoder stages. We address this with a \emph{dual-branch decoder block} that processes both aspects in parallel---a depth-wise convolutional boundary branch and a large-kernel-attention context branch---and fuses them adaptively.

\subsubsection{Architecture}

Given input features $\hat{\mathbf{D}}_s$ (\cref{fig:biref}(b)), we first apply a $1\times1$ convolution with LayerNorm and GELU to reduce channels to $64$. The reduced features $\mathbf{X}$ are then processed by two~parallel branches:

\textbf{Boundary branch.}
Two stacked depth-wise $3\times3$ convolutions with LayerNorm capture local spatial patterns and high-frequency boundary cues with minimal parameters:
\begin{equation}
    \mathbf{B} = \text{DWConv}_{3\times3}(\text{DWConv}_{3\times3}(\mathbf{X})),
\end{equation}
where each depth-wise convolution is followed by LayerNorm, with a GELU between the two.

\textbf{Context branch.}
The context branch aggregates structural context through \emph{large-kernel attention} (LKA)~\cite{VAN}, which emulates a large kernel (a $23\times23$ effective receptive field) by a decomposed convolution chain---a $5\times5$ depth-wise convolution, a $7\times7$ depth-wise convolution with dilation $3$, and a $1\times1$ point-wise convolution---whose response acts as an element-wise attention map:
\begin{equation}
    \text{LKA}(\mathbf{Z}) = \mathbf{Z} \odot \text{Conv}_{1\times1}\big(\text{DWConv}_{7\times7}^{d=3}(\text{DWConv}_{5\times5}(\mathbf{Z}))\big).
\end{equation}
This data-dependent long-range interaction costs only $O(N)$. The branch follows a pre-norm layout: the reduced features are normalized by LayerNorm, projected by a $1\times1$ convolution with GELU, attended by LKA, projected back by a second $1\times1$ convolution, and added residually; a feed-forward network (expansion ratio $2$) with LayerNorm completes the branch, yielding the context features $\mathbf{C}$.

\subsubsection{Adaptive Fusion}

The two branches are fused via a learned gate that adaptively weighs boundary and context features:
\begin{equation}
    \mathbf{O} = \mathbf{g}_f \odot \mathbf{B} + (1 - \mathbf{g}_f) \odot \mathbf{C} + \beta \cdot \text{Proj}([\mathbf{B}; \mathbf{C}]),
\end{equation}
where $\mathbf{g}_f = \sigma(\text{Conv}_{1\times1}([\mathbf{B}; \mathbf{C}]))$ is a fusion gate with one value per channel and pixel (distinct from the SGG gating signal $\mathbf{g}$), $\text{Proj}$ is a $1\times1$ projection, and $\beta=0.1$ is a fixed scaling factor for the cross-branch residual. The fused features pass through a $3\times3$ convolution with LayerNorm that restores the stage width, and are summed with a shortcut that projects the block input $\hat{\mathbf{D}}_s$ to the same width by a $1\times1$ convolution with LayerNorm; the sum is the block output $\mathbf{Y}_s \in \mathbb{R}^{C_{s+1}\times h\times w}$, which the outward reference (\cref{sec:outward}) refines before it is upsampled and added to the next encoder feature (at $s_0$--$s_2$).

\subsection{Outward Reference}
\label{sec:outward}

While \ourmodelVOne{}~\cite{BiRefNet} computes a Laplacian edge map from the input image and uses single-channel attention to weight decoder features ($\mathbf{Y}_s \leftarrow \mathbf{Y}_s \odot a$, $a\in\mathbb{R}^{1\times h\times w}$), our redesign (\cref{fig:biref}(c)) introduces three improvements: prediction-gated image-edge supervision, lightweight boundary feature extraction, and full-channel residual attention.

\subsubsection{Prediction-Gated Image-Edge Supervision}

\ourmodelVOne{} already multiplies its Laplacian target by its intermediate prediction, but does so in logit space---a signed Laplacian times an unbounded logit, squashed by a sigmoid only at the loss (\cref{sec:introduction}). We instead build the target in two steps, bounding the edge map and the gate separately. First, an edge map is extracted from the input image by a lightly smoothed Laplacian: the grayscale input is filtered by a $5\times5$ bilateral filter, followed by a $5\times5$ Laplacian whose absolute response is squashed by $\tanh$:
\begin{equation}
    \mathbf{E} = \tanh\big(\lvert\text{Lap}_{5}(\text{Bilateral}_{5}(\text{gray}(\mathbf{I})))\rvert\big).
\end{equation}
Second, the supervision target for stage $s$ is constructed by multiplying $\mathbf{E}$ with the intermediate sigmoid prediction $\hat{\mathbf{M}}_s$, confining the boundary learning to the edges inside the object currently being segmented while discarding background clutter edges:
\begin{equation}
    \mathbf{G}_s^{\text{label}} = \mathbf{E} \odot \hat{\mathbf{M}}_s,
\end{equation}
where the supervision is computed at the input resolution for every stage. The target is thus available without any ground-truth boundary annotation: the auxiliary supervision asks the model to respect exactly the visible edge evidence inside its own predicted foreground. No stop-gradient is applied to the gating prediction $\hat{\mathbf{M}}_s$, so the boundary loss also propagates through the gate into the segmentation head, reaching the mask prediction itself: the boundary and segmentation streams are thus optimized jointly rather than in isolation.

\subsubsection{Lightweight Boundary Feature Extraction}

From the block output $\mathbf{Y}_s$ at stage $s$, a depth-wise separable block extracts boundary features: a $3\times3$ depth-wise convolution captures local spatial patterns, and a $1\times1$ point-wise convolution projects them to a compact $16$-channel representation, each followed by LayerNorm and GELU:
\begin{equation}
    \mathbf{H}_s = \text{PWConv}_{1\times1}(\text{DWConv}_{3\times3}(\mathbf{Y}_s)).
\end{equation}
From $\mathbf{H}_s$, we predict the boundary map $\hat{\mathbf{G}}_s = \sigma(\text{Conv}_{1\times1}(\mathbf{H}_s))$, which is supervised against $\mathbf{G}_s^{\text{label}}$ during training. Boundary evidence is a local, low-level signal, so a depth-wise separable block suffices to isolate it; keeping the stream compact ensures the auxiliary path refines, rather than competes with, the main decoding pathway---the entire outward-reference stack adds only ${\sim}0.06$M parameters across the three decoder stages it serves.

\subsubsection{Full-Channel Residual Attention}

The key refinement mechanism is a channel-wise attention map $\mathbf{A}_s \in \mathbb{R}^{C_{s+1} \times h \times w}$ (the shape of $\mathbf{Y}_s$: 768, 384, and 192 channels at $s_0$--$s_2$), produced from the boundary features by a second, separate $1\times1$ convolution head:
\begin{equation}
    \mathbf{A}_s = \sigma(\text{Conv}_{1\times1}(\mathbf{H}_s)),
\end{equation}
modulating the block output in place via residual attention:
\begin{equation}
    \mathbf{Y}_s \leftarrow \mathbf{Y}_s \odot (1 + \mathbf{A}_s).
\end{equation}
This formulation offers two advantages over \ourmodelVOne{}'s $\mathbf{Y}_s \odot a$: (1) the full-channel attention enables per-channel modulation, selectively amplifying each feature dimension by its own factor in $(1,2)$ instead of scaling all channels by the same factor; (2) the residual form $(1+\mathbf{A}_s)$ preserves the original features when $\mathbf{A}_s \approx 0$, preventing information loss and stabilizing gradients.

\subsection{Training and Inference}
\label{sec:training}

\subsubsection{Training Procedure}

\textbf{Backbone.}
We keep the DINOv3-L backbone frozen except for its last transformer stage, which is fine-tuned at $0.05\times$ the base learning rate, so that adapting the pretrained representation refines rather than overwrites it. Together with the decoder, squeeze module, pyramid neck, and the inward- and outward-reference modules, this gives 84.9M trainable parameters. \Cref{sec:ablation} quantifies this partial adaptation on both sides: it clearly outperforms a fully frozen backbone, while tuning deeper---the last two stages---is both costlier and~less~accurate.

\textbf{Normalization and activation.}
We replace BatchNorm and ReLU with LayerNorm and GELU throughout the decoder: LayerNorm operates independently of batch statistics and remains stable under bf16 mixed precision, matching the ViT backbone's convention.

\textbf{Optimization.}
We train with the Lion optimizer~\cite{Lion} under a linear-warmup--cosine schedule; the hyperparameter configuration is listed in \cref{sec:impl}.

\textbf{Loss function.}
The total loss combines pixel-level losses and boundary prediction losses:
\begin{equation}
    \mathcal{L} = \mathcal{L}_\text{pix} + \mathcal{L}_\text{bdr},
\end{equation}
where $\mathcal{L}_\text{pix}$ applies BCE and IoU to every prediction in $\mathcal{P}=\{\hat{\mathbf{M}}_s\}\cup\{\hat{\mathbf{M}}\}$ against the ground-truth mask $\mathbf{M}$, and SSIM~\cite{SSIM} to the final output only:
\begin{equation}
    \begin{aligned}
    \mathcal{L}_\text{pix}
    =&
    \sum_{\hat{\mathbf{M}}_s \in \mathcal{P}}
    \left(
        \lambda_\text{bce}\mathcal{L}_\text{BCE}(\hat{\mathbf{M}}_s,\mathbf{M})
        + \lambda_\text{iou}\mathcal{L}_\text{IoU}(\hat{\mathbf{M}}_s,\mathbf{M})
    \right) \\
    &+
    \lambda_\text{ssim}\mathcal{L}_\text{SSIM}(\hat{\mathbf{M}},\mathbf{M}) ,
    \end{aligned}
\end{equation}
with loss weights given in \cref{sec:impl}. The boundary loss $\mathcal{L}_\text{bdr}$ supervises the boundary predictions of the outward-reference modules against $\mathbf{G}_s^{\text{label}}$ using BCE loss.

\subsubsection{Inference Procedure}
\label{sec:inference}

During inference, \ourmodel{} performs a single forward pass without the auxiliary boundary supervision; the outward reference still contributes through its learned attention map $\mathbf{A}_s$.

\section{Experiments}
\label{sec:experiment}

We evaluate \ourmodel{} on three high-precision segmentation tasks: DIS, high-resolution salient object detection (HRSOD), and camouflaged object detection (COD). All three tasks require pixel-accurate segmentation masks with fine structural details.

\subsection{Datasets and Settings}

\textbf{DIS.}
We use the DIS5K benchmark~\cite{DIS5K}, which contains high-resolution images across 225 categories with pixel-accurate annotations. The training set (DIS-TR) comprises 3,000 images. We report results on five evaluation sets: DIS-VD (470 images), DIS-TE1 (500), DIS-TE2 (500), DIS-TE3 (500), and DIS-TE4 (500), the four DIS-TE splits being organized by increasing structural complexity.

\textbf{HRSOD.}
Following~\cite{PGNet,InSPyReNet,BiRefNet}, we adopt different combinations of the training splits of DUTS~\cite{DUTS} (10,553 images), HRSOD~\cite{HRSOD} (1,610 images), and UHRSD~\cite{PGNet} (4,932 images) as the training set. Evaluation is conducted on three high-resolution benchmarks: DAVIS-S~\cite{HRSOD} (92 images), HRSOD-TE~\cite{HRSOD} (400 images), and UHRSD-TE~\cite{PGNet} (988 images). We additionally report the two standard low-resolution salient object detection test sets, DUTS-TE~\cite{DUTS} (5,019 images) and DUT-OMRON~\cite{DUT-OMRON} (5,168 images), evaluated with the same checkpoints without any re-training.

\textbf{COD.}
Following the standard protocol~\cite{SINet_v2}, we train on the combined training splits of COD10K~\cite{SINet_v1} (3,040 images) and CAMO~\cite{CAMO} (1,000 images). Evaluation is conducted on three test sets: CAMO-TE~\cite{CAMO} (250 images), COD10K-TE~\cite{SINet_v1} (2,026 images), and NC4K~\cite{SLSR} (4,121 images).

\begin{table*}[t!]
    \setlength{\belowcaptionskip}{0cm}
    \renewcommand{\arraystretch}{1.0}
    \setlength{\tabcolsep}{1.98pt}
    \footnotesize
    \centering
    \caption{\textbf{Quantitative comparison on DIS5K.} All methods use $1024\times1024$ input; DIS-TEs is the mean of the per-split scores, and best results are in \textbf{bold}. $^{\dagger}$We recompute GenPercept's scores from its official inference, changing only the input resolution from its native 768. $^{\ddagger}$Unlike the other methods, DiffDIS is evaluated on fully binarized maps, following its official protocol. Hyperlinked method names link to their released~code}
    \label{tab:sota_dis}
    \vspace{-10pt}
    {\hypersetup{urlcolor=cvprblue}\begin{tabular}{lcccccc|cccccc|cccccc}
        \toprule
        \multirow{2}{*}{\textbf{Methods}} & \multicolumn{6}{c|}{\textbf{DIS-TE1 (500)}}  & \multicolumn{6}{c|}{\textbf{DIS-TE2 (500)}}& \multicolumn{6}{c}{\textbf{DIS-TE3 (500)}}\\
        \cmidrule[0.05em](lr){2-7} \cmidrule[0.05em](lr){8-13} \cmidrule[0.05em](lr){14-19}
        \metricsDIS{}\metricsDIS{}\metricsDIS{} \\
        \midrule
        \href{https://github.com/xuebinqin/DIS}{IS-Net$_{22}$}~\cite{DIS5K} & .740 & .662 & .074 & .787 & .820 & 149 & .799 & .728 & .070 & .823 & .858 & 340 & .830 & .758 & .064 & .836 & .883 & 687 \\
        \href{https://github.com/plemeri/InSPyReNet}{InSPyReNet$_{22}$}~\cite{InSPyReNet} & .845 & .788 & .043 & .873 & .894 & 110 & .894 & .846 & .036 & .905 & .928 & 255 & .919 & .871 & .034 & .918 & .943 & 522 \\
        FP-DIS$_{23}$~\cite{FP-DIS} & .784 & .713 & .060 & .821 & .860 & 160 & .827 & .767 & .059 & .845 & .893 & 373 & .868 & .811 & .049 & .871 & .922 & 780 \\
        \href{https://github.com/PJLallen/UDUN}{UDUN$_{23}$}~\cite{UDUN} & .784 & .720 & .059 & .817 & .864 & 140 & .829 & .768 & .058 & .843 & .886 & 325 & .865 & .809 & .050 & .865 & .917 & 658 \\
        \href{https://github.com/qianyu-dlut/MVANet}{MVANet$_{24}$}~\cite{MVANet} & .873 & .823 & .037 & .879 & .916 & 105 & .916 & .874 & .030 & .915 & .943 & 252 & .929 & .890 & .031 & .920 & .952 & 527 \\
        \href{https://github.com/aim-uofa/GenPercept}{GenPercept$_{25}^{\dagger}$}~\cite{GenPercept} & .839 & .818 & .036 & .873 & .917 & 215 & .874 & .854 & .034 & .887 & .937 & 503 & .892 & .876 & .032 & .894 & .954 & 965 \\
        \href{https://github.com/qianyu-dlut/DiffDIS}{DiffDIS$_{25}^{\ddagger}$}~\cite{DiffDIS} & .883 & .862 & .030 & .891 & .933 & 104 & .917 & .895 & .026 & .913 & .951 & 267 & .934 & .916 & .025 & .919 & .964 & 595 \\
        \href{https://github.com/Tennine2077/PDFNet}{PDFNet$_{26}$}~\cite{PDFNet} & .891 & .848 & .030 & .899 & .928 & 114 & .920 & .886 & .027 & .924 & .948 & 273 & .936 & .901 & .027 & .928 & .957 & 576 \\
        \href{https://github.com/ZhengPeng7/BiRefNet}{\ourmodelVOne{}$_{24}$}~\cite{BiRefNet} & .860 & .819 & .037 & .885 & .911 & 106 & .894 & .857 & .036 & .900 & .930 & 266 & .925 & .893 & .028 & .919 & .955 & 569 \\
        \rowcolor{mygray}
        \textbf{\ourmodel{}} & \textbf{.914} & \textbf{.890} & \textbf{.022} & \textbf{.927} & \textbf{.949} & \textbf{94} & \textbf{.939} & \textbf{.915} & \textbf{.020} & \textbf{.940} & \textbf{.960} & \textbf{225} & \textbf{.951} & \textbf{.926} & \textbf{.019} & \textbf{.942} & \textbf{.968} & \textbf{473} \\
        \midrule
        \multirow{2}{*}{\textbf{Methods}} & \multicolumn{6}{c|}{\textbf{DIS-TE4 (500)}}  & \multicolumn{6}{c|}{\textbf{DIS-TEs (TE1--TE4) (2,000)}}& \multicolumn{6}{c}{\textbf{DIS-VD (470)}}\\
        \cmidrule[0.05em](lr){2-7} \cmidrule[0.05em](lr){8-13} \cmidrule[0.05em](lr){14-19}
        \metricsDIS{}\metricsDIS{}\metricsDIS{} \\
        \midrule
        \href{https://github.com/xuebinqin/DIS}{IS-Net$_{22}$}~\cite{DIS5K} & .827 & .753 & .072 & .830 & .870 & 2888 & .799 & .726 & .070 & .819 & .858 & 1016 & .791 & .717 & .074 & .813 & .856 & 1116 \\
        \href{https://github.com/plemeri/InSPyReNet}{InSPyReNet$_{22}$}~\cite{InSPyReNet} & .905 & .848 & .042 & .905 & .928 & 2336 & .891 & .838 & .039 & .900 & .923 & 806 & .889 & .834 & .042 & .900 & .922 & 904 \\
        FP-DIS$_{23}$~\cite{FP-DIS} & .846 & .788 & .061 & .852 & .906 & 3347 & .831 & .770 & .057 & .847 & .895 & 1165 & .823 & .763 & .062 & .843 & .891 & 1309 \\
        \href{https://github.com/PJLallen/UDUN}{UDUN$_{23}$}~\cite{UDUN} & .846 & .792 & .059 & .849 & .901 & 2785 & .831 & .772 & .057 & .844 & .892 & 977 & .823 & .763 & .059 & .838 & .892 & 1097 \\
        \href{https://github.com/qianyu-dlut/MVANet}{MVANet$_{24}$}~\cite{MVANet} & .912 & .857 & .041 & .903 & .938 & 2347 & .908 & .861 & .035 & .904 & .937 & 808 & .904 & .863 & .034 & .905 & .936 & 899 \\
        \href{https://github.com/aim-uofa/GenPercept}{GenPercept$_{25}^{\dagger}$}~\cite{GenPercept} & .864 & .850 & .043 & .868 & .944 & 3769 & .867 & .850 & .037 & .881 & .938 & 1363 & .868 & .852 & .037 & .882 & .938 & 1434 \\
        \href{https://github.com/qianyu-dlut/DiffDIS}{DiffDIS$_{25}^{\ddagger}$}~\cite{DiffDIS} & .909 & .893 & .032 & .896 & .955 & 2762 & .911 & .892 & .028 & .905 & .950 & 932 & .908 & .888 & .029 & .904 & .948 & 1055 \\
        \href{https://github.com/Tennine2077/PDFNet}{PDFNet$_{26}$}~\cite{PDFNet} & .912 & .869 & .037 & .910 & .941 & 2510 & .915 & .876 & .030 & .916 & .944 & 868 & .915 & .875 & .029 & .916 & .945 & 946 \\
        \href{https://github.com/ZhengPeng7/BiRefNet}{\ourmodelVOne{}$_{24}$}~\cite{BiRefNet} & .904 & .864 & .039 & .900 & .939 & 2723 & .896 & .858 & .035 & .901 & .934 & 916 & .891 & .854 & .038 & .898 & .931 & 989 \\
        \rowcolor{mygray}
        \textbf{\ourmodel{}} & \textbf{.939} & \textbf{.908} & \textbf{.027} & \textbf{.933} & \textbf{.959} & \textbf{2150} & \textbf{.936} & \textbf{.910} & \textbf{.022} & \textbf{.936} & \textbf{.959} & \textbf{736} & \textbf{.932} & \textbf{.907} & \textbf{.024} & \textbf{.935} & \textbf{.956} & \textbf{731} \\
        \bottomrule
    \end{tabular}}
\end{table*}

\subsection{Evaluation Metrics}

For DIS, following~\cite{DIS5K,UDUN,BiRefNet}, we adopt six widely used metrics: maximum F-measure ($F_{\beta}^{x}$)~\cite{Fmeasure}, weighted F-measure ($F_{\beta}^{\omega}$)~\cite{WFmeasure}, mean absolute error ($\mathcal{M}$), structure measure ($S_m$)~\cite{Smeasure}, mean E-measure ($E_{\phi}^{m}$)~\cite{Emeasure}, and relaxed human correction efforts ($HCE_{\gamma}$)~\cite{DIS5K}, the last of which approximates the human effort (\ie, mouse clicks within tolerance $\gamma{=}5$) required to correct faulty predictions.
For HRSOD, following~\cite{PGNet,BiRefNet}, we report $S_m$, $F_{\beta}^{x}$, $E_{\phi}^{m}$, and $\mathcal{M}$.
For COD, following~\cite{SINet_v2,BiRefNet}, we report $S_m$, $F_{\beta}^{\omega}$, mean F-measure ($F_{\beta}^{m}$), $E_{\phi}^{m}$, max E-measure ($E_{\phi}^{x}$), and $\mathcal{M}$.

\subsection{Implementation Details}
\label{sec:impl}

We implement \ourmodel{} using PyTorch~\cite{PyTorch}. Training uses two GPUs. Unless otherwise stated, all measurements are performed on a single NVIDIA GeForce RTX 5090.
Backbone freezing and the DINOv3 last-stage tuning follow \cref{sec:training}; all decoder-side components are optimized at the base learning rate. The base learning rate is $3\times10^{-4}$, and the total batch size is $8$, split evenly across the training GPUs. We apply gradient clipping at norm $10.0$. Training uses bf16 mixed precision with a compiled computation graph. Unless otherwise stated, input images are resized to $1024\times1024$ for both training and inference. We use random horizontal flipping and foreground--background recomposition for data augmentation.

The model is trained for 50 epochs on DIS5K, 30 epochs on COD, and 20 epochs on HRSOD. The loss function applies BCE ($\lambda{=}30$) and IoU ($\lambda{=}2$) losses to all decoder predictions, applies SSIM ($\lambda{=}400$) to the final prediction, and adds a BCE loss for the outward-reference boundary predictions; the weights are set so that the three pixel-level terms stay within the same order of magnitude. In the state-of-the-art comparisons (\cref{tab:sota_dis,tab:sota_hrsod,tab:sota_cod} and \cref{fig:sota_radar}), every number reported for a training configuration comes from one checkpoint, with no per-set selection: the checkpoint that maximizes the task's primary metric---$F_{\beta}^{x}$ for DIS, $S_m$ for HRSOD and COD---averaged with equal weight over its primary evaluation sets, namely DIS-VD and the four DIS-TE splits for DIS; CAMO-TE, COD10K-TE, and NC4K for COD; and DAVIS-S, HRSOD-TE, and UHRSD-TE for HRSOD. The two low-resolution sets of \cref{tab:sota_hrsod} are reported without taking part in the selection. The output masks are resized back to the original resolution via bilinear interpolation for evaluation. Detailed model size, GMACs, and inference speed analyses are reported in~\cref{sec:efficiency}. Pretrained models\footnote{{\url{https://huggingface.co/collections/ZhengPeng7/birefnet-v2}}} and an interactive online demo\footnote{{\url{https://huggingface.co/spaces/ZhengPeng7/BiRefNet_demo}}} have been released.

\subsection{Comparison with State-of-the-Art Methods}

\subsubsection{DIS5K}

As shown in~\cref{tab:sota_dis}, \ourmodel{} achieves the best score in every metric on all five evaluation sets. Compared with PDFNet~\cite{PDFNet}---the strongest competitor in the table on the primary metric, which additionally consumes a pseudo-depth map produced by a 335\,M monocular estimator---on DIS-VD $F_{\beta}^{x}$ improves from 0.915 to 0.932 ($+$1.7 points), $S_m$ from 0.916 to 0.935, $F_{\beta}^{\omega}$ from 0.875 to 0.907, and $HCE_{\gamma}$ falls from 946 to 731. Against DiffDIS~\cite{DiffDIS}, the strongest competitor that, like ours, takes nothing beyond the RGB image, the margins widen to $+$2.4 points of $F_{\beta}^{x}$ (0.908 to 0.932) and $+$3.1 of $S_m$ (0.904 to 0.935), with $\mathcal{M}$ dropping from 0.029 to 0.024. \ourmodel{} consistently achieves the lowest $HCE_{\gamma}$; the advantage is most pronounced on the hardest split, DIS-TE4, where \ourmodel{} lowers the correction effort to 2150---below the previous best of 2336 (InSPyReNet) and $\sim$21\% below \ourmodelVOne{} (2723).

\subsubsection{HRSOD}

\begin{table*}[t!]
    \setlength{\belowcaptionskip}{0cm}
    \renewcommand{\arraystretch}{1.0}
    \setlength{\tabcolsep}{2.12pt}
    \scriptsize
    \centering
    \caption{\textbf{Quantitative comparison on high- and low-resolution salient object detection benchmarks.} The last two benchmarks are low-resolution. TR is the training set: 1, 2, and 3 denote DUTS, HRSOD, and UHRSD. Second-best results are \underline{underlined}, and ``-'' marks unavailable scores; $^{\P}$whole row taken from \cite{PGNet}, which reports more of these scores than the method's own paper}
    \label{tab:sota_hrsod}
    \vspace{-10pt}
    {\hypersetup{urlcolor=cvprblue}\begin{tabular}{lc|cccc|cccc|cccc|cccc|cccc}
        \toprule
        \multirow{2}{*}{\textbf{Methods}} & \multirow{2}{*}{\textbf{TR}} & \multicolumn{4}{c|}{\textbf{DAVIS-S (92)}} & \multicolumn{4}{c|}{\textbf{HRSOD-TE (400)}} & \multicolumn{4}{c|}{\textbf{UHRSD-TE (988)}} & \multicolumn{4}{c|}{\textbf{DUTS-TE (5,019)}} & \multicolumn{4}{c}{\textbf{DUT-OMRON (5,168)}} \\
        \cmidrule[0.05em](lr){3-6} \cmidrule[0.05em](lr){7-10} \cmidrule[0.05em](lr){11-14} \cmidrule[0.05em](lr){15-18} \cmidrule[0.05em](lr){19-22}
        & \metricsSODHR{}\metricsSODHR{}\metricsSODHR{}\metricsSODLR{}\metricsSODLR{} \\
        \midrule
        \href{https://github.com/yi94code/HRSOD}{HRSOD$_{19}^{\P}$}~\cite{HRSOD}       & 1,2     & .876 & .899 & .955 & .026 & .896 & .905 & .934 & .030 & -    & -    & -    & -    & .824 & .835 & .885 & .050 & .762 & .743 & .831 & .065 \\
        \href{https://github.com/weijun-arc/LDF}{LDF$_{20}^{\P}$}~\cite{LDF}           & 1       & .922 & .911 & .947 & .019 & .904 & .904 & .919 & .032 & .888 & .913 & .891 & .047 & .892 & .898 & .910 & .034 & .838 & .820 & .873 & .051 \\
        DHQ$_{21}$~\cite{DHQSOD}        & 1,2     & .939 & .938 & -    & .012 & .922 & .922 & -    & .022 & -    & -    & -    & -    & .892 & .900 & -    & .031 & .843 & .820 & -    & .048 \\
        \href{https://github.com/iCVTEAM/PGNet}{PGNet$_{22}$}~\cite{PGNet}       & 2,3     & .954 & .957 & .979 & .010 & .938 & .945 & .946 & .020 & .935 & .949 & .916 & .026 & .859 & .871 & .897 & .038 & .786 & .772 & .884 & .058 \\
        \href{https://github.com/ZhengPeng7/BiRefNet}{\ourmodelVOne{}$_{24}$}~\cite{BiRefNet} & 1       & .967 & .966 & .984 & .008 & .957 & .958 & .972 & .014 & .931 & .933 & .943 & .030 & .939 & .937 & .958 & .019 & .868 & .813 & .878 & .040 \\
        \href{https://github.com/ZhengPeng7/BiRefNet}{\ourmodelVOne{}$_{24}$}~\cite{BiRefNet} & 2,3     & .976 & .980 & \underline{.990} & \underline{.006} & .956 & .953 & .967 & .016 & .952 & .958 & .964 & .019 & .933 & .928 & .954 & .020 & .864 & .810 & .879 & .040 \\
        \href{https://github.com/ZhengPeng7/BiRefNet}{\ourmodelVOne{}$_{24}$}~\cite{BiRefNet} & 1,2,3   & .975 & .979 & .989 & \underline{.006} & \underline{.962} & \underline{.961} & .973 & .013 & .957 & .963 & .969 & \underline{.016} & \underline{.944} & \underline{.943} & .962 & \underline{.018} & .882 & .839 & .896 & .038 \\
        \rowcolor{mygray}
        \textbf{\ourmodel{}}            & 1       & .974 & .975 & .989 & \underline{.006} & \underline{.962} & \underline{.961} & \underline{.976} & \underline{.012} & .943 & .949 & .956 & .025 & .940 & .933 & .957 & .020 & .876 & .834 & .895 & .043 \\
        \rowcolor{mygray}
        \textbf{\ourmodel{}}            & 2,3     & \textbf{.980} & \textbf{.985} & \textbf{.993} & \textbf{.004} & \textbf{.968} & \textbf{.969} & \textbf{.980} & \textbf{.010} & \underline{.959} & \underline{.966} & \underline{.971} & \underline{.016} & .942 & \underline{.943} & \underline{.964} & \underline{.018} & \underline{.890} & \underline{.861} & \textbf{.918} & \underline{.036} \\
        \rowcolor{mygray}
        \textbf{\ourmodel{}}            & 1,2,3   & \underline{.979} & \underline{.984} & \textbf{.993} & \textbf{.004} & \textbf{.968} & \textbf{.969} & \textbf{.980} & \textbf{.010} & \textbf{.960} & \textbf{.967} & \textbf{.973} & \textbf{.015} & \textbf{.950} & \textbf{.949} & \textbf{.969} & \textbf{.016} & \textbf{.895} & \textbf{.864} & \underline{.916} & \textbf{.035} \\
        \bottomrule
    \end{tabular}}
\end{table*}

\cref{tab:sota_hrsod} reports results on three high-resolution and two low-resolution salient object detection benchmarks, comparing methods under matched training data (the TR column). Under the identical TR=2,3 setting, \ourmodel{} surpasses \ourmodelVOne{} on every metric of all three benchmarks (\eg, HRSOD-TE $S_m$ 0.968 vs 0.956 and $\mathcal{M}$ 0.010 vs 0.016), achieving the strongest results among methods trained on matched data. Adding the low-resolution DUTS set on top (TR=1,2,3) performs on par with TR=2,3---the two agree to 0.969 in $S_m$ averaged over the three benchmarks, and no single benchmark separates them by more than 0.001 in $S_m$---so on these benchmarks the two high-resolution sets already saturate the benefit of the extra DUTS data. Even trained on DUTS alone (TR=1), \ourmodel{} outperforms the identically trained LDF by large margins (\eg, $+$6.4 points of $F_{\beta}^{x}$ on DAVIS-S) and surpasses \ourmodelVOne{} under the same TR=1 setting on every metric of all three benchmarks, indicating that the gain does not stem from high-resolution training data. The two low-resolution test sets probe the converse risk---that an architecture built around a high-resolution detail stream might pay for it at conventional resolutions---and the same checkpoints take the best value of every metric on both: under identical training data (TR=1,2,3) \ourmodel{} improves on \ourmodelVOne{} by $+$0.6 points of $S_m$ on DUTS-TE and by $+$1.3 on DUT-OMRON---matching or exceeding the gain on any of the three high-resolution benchmarks. The detail-oriented redesign therefore costs nothing where the input carries little detail to begin with.

\subsubsection{COD}

\begin{table*}[t!]
    \setlength{\belowcaptionskip}{0cm}
    \renewcommand{\arraystretch}{1.0}
    \setlength{\tabcolsep}{2.80pt}
    \footnotesize
    \centering
    \caption{\textbf{Quantitative comparison on COD benchmarks.} Competitors' scores are from their papers where the metric is reported with the same definition; $^{\dagger}$remaining cells recomputed from the released predictions of the same model; $^{\ddagger}$whole row recomputed from the released model, which is not the cited version's checkpoint but scores within 0.01 of it on every metric both report; $^{\P}$whole row taken from re-evaluations of the released predictions reported in \cite{BiRefNet} (SINet: the authors' retrained release, as reported in \cite{ZoomNeXt,ESCNet}), as the cited papers report fewer metrics}
    \label{tab:sota_cod}
    {\hypersetup{urlcolor=cvprblue}\begin{tabular}{lcccccc|cccccc|cccccc}
        \toprule
        \multirow{2}{*}{\textbf{Methods}} & \multicolumn{6}{c|}{\textbf{CAMO-TE (250)}} & \multicolumn{6}{c|}{\textbf{COD10K-TE (2,026)}} & \multicolumn{6}{c}{\textbf{NC4K (4,121)}} \\
        \cmidrule[0.05em](lr){2-7} \cmidrule[0.05em](lr){8-13} \cmidrule[0.05em](lr){14-19}
        \metricsCOD{}\metricsCOD{}\metricsCOD{} \\
        \midrule
        \href{https://github.com/DengPingFan/SINet}{SINet$_{20}^{\P}$}~\cite{SINet_v1} & .745 & .644 & .702 & .804 & .829 & .091 & .776 & .631 & .679 & .864 & .874 & .043 & .808 & .723 & .769 & .871 & .883 & .058 \\
        \href{https://github.com/GewelsJI/SINet-V2}{SINet-V2$_{22}^{\dagger}$}~\cite{SINet_v2} & .820 & .743 & .782 & .882 & .895 & .070 & .815 & .680 & .718 & .887 & .906 & .037 & .847 & .770 & .805 & .903 & .914 & .048 \\
        \href{https://github.com/ChunmingHe/FEDER}{FEDER$_{23}^{\P}$}~\cite{FEDER} & .802 & .738 & .781 & .867 & .873 & .071 & .822 & .716 & .751 & .900 & .905 & .032 & .847 & .789 & .824 & .907 & .915 & .044 \\
        \href{https://github.com/HUuxiaobin/HitNet}{HitNet$_{23}^{\P}$}~\cite{HitNet} & .849 & .809 & .831 & .906 & .910 & .055 & .871 & .806 & .823 & .935 & .938 & .023 & .875 & .834 & .853 & .926 & .929 & .037 \\
        \href{https://github.com/ZhouHuang23/FSPNet}{FSPNet$_{23}$}~\cite{FSPNet} & .856 & .799 & .830 & .899 & .928 & .050 & .851 & .735 & .769 & .895 & .930 & .026 & .879 & .816 & .843 & .915 & .937 & .035 \\
        \href{https://github.com/HVision-NKU/CamoFormer}{CamoFormer$_{24}^{\ddagger}$}~\cite{CamoFormer} & .872 & .831 & .854 & .929 & .938 & .046 & .869 & .786 & .811 & .932 & .939 & .023 & .892 & .847 & .868 & .939 & .946 & .030 \\
        \href{https://github.com/lartpang/ZoomNeXt}{ZoomNeXt$_{24}^{\dagger}$}~\cite{ZoomNeXt} & .889 & .857 & .875 & .935 & .945 & .041 & .898 & .827 & .848 & .941 & .956 & .018 & .903 & .863 & .884 & .940 & .951 & .028 \\
        \href{https://github.com/CSYSI/FSEL}{FSEL$_{24}^{\dagger}$}~\cite{FSEL} & .885 & .851 & .868 & .942 & .948 & .040 & .873 & .800 & .817 & .937 & .947 & .021 & .892 & .853 & .871 & .942 & .947 & .030 \\
        \href{https://github.com/suy9/ESCNet}{ESCNet$_{25}^{\ddagger}$}~\cite{ESCNet} & .875 & .849 & .867 & .937 & .940 & .041 & .873 & .808 & .827 & .942 & .946 & .020 & .893 & .864 & .881 & .945 & .949 & .028 \\
        \href{https://github.com/ChunmingHe/RUN}{RUN$_{25}^{\ddagger}$}~\cite{RUN} & .804 & .739 & .779 & .864 & .873 & .070 & .823 & .717 & .751 & .898 & .904 & .031 & .850 & .790 & .822 & .908 & .915 & .043 \\
        \href{https://github.com/Sssssuperior/VSCode}{VSCode-v2$_{26}^{\dagger}$}~\cite{VSCode-v2} & .875 & .817 & .840 & .919 & .932 & .049 & .874 & .787 & .811 & .930 & .942 & .024 & .897 & .845 & .866 & .934 & .944 & .031 \\
        \href{https://github.com/ZhengPeng7/BiRefNet}{\ourmodelVOne{}$_{24}$}~\cite{BiRefNet} & \underline{.904} & \underline{.890} & \underline{.904} & \underline{.954} & \underline{.959} & \underline{.030} & \underline{.913} & \underline{.874} & \underline{.888} & \underline{.960} & \underline{.967} & \underline{.014} & \underline{.914} & \underline{.894} & \underline{.909} & \underline{.953} & \underline{.960} & \underline{.023} \\
        \rowcolor{mygray}
        \textbf{\ourmodel{}} & \textbf{.925} & \textbf{.909} & \textbf{.916} & \textbf{.963} & \textbf{.971} & \textbf{.024} & \textbf{.934} & \textbf{.901} & \textbf{.906} & \textbf{.970} & \textbf{.978} & \textbf{.012} & \textbf{.928} & \textbf{.908} & \textbf{.918} & \textbf{.959} & \textbf{.967} & \textbf{.020} \\
        \bottomrule
    \end{tabular}}
\end{table*}

\begin{figure}[t!]
\centering
\includegraphics[width=.8\linewidth]{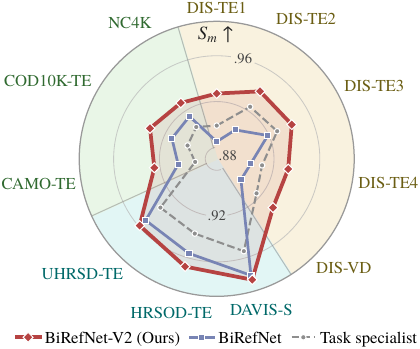}
\caption{\textbf{Cross-task architectural transferability.} Sectors: DIS (yellow), HRSOD (teal; TR=2,3), and COD (green); task specialists (strongest compared): \emph{PDFNet}, \emph{PGNet}, and \emph{ZoomNeXt}, respectively. The radial axis runs from $0.87$ (center) to $0.99$ (rim), which enlarges the differences.}
\label{fig:sota_radar}
\end{figure}

\begin{figure*}[t]
\centering
\includegraphics[width=\linewidth]{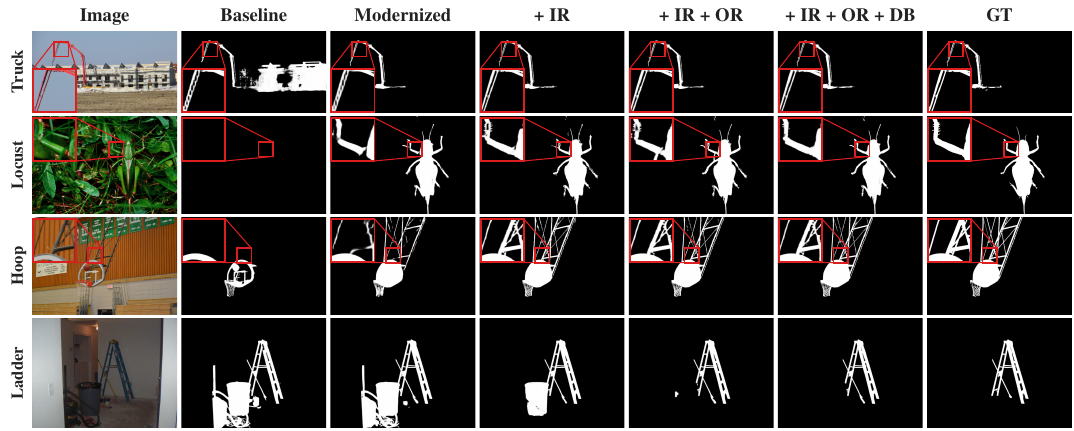}
\vspace{-20pt}
\caption{\textbf{The cumulative component build-up, visualized.} Insets (top three rows) enlarge the red-boxed region.}
\label{fig:ablation_visual}
\end{figure*}

As shown in~\cref{tab:sota_cod}, \ourmodel{} establishes a new state of the art on all three COD benchmarks across every metric. On NC4K (the largest test set), it surpasses the strongest baseline \ourmodelVOne{} by $+$1.4 points on both $S_m$ and $F_{\beta}^{\omega}$, with a $\sim$13\% relative reduction in $\mathcal{M}$. The gain over recent specialist COD models is even larger: \ourmodel{} outperforms ZoomNeXt~\cite{ZoomNeXt}, FSEL~\cite{FSEL}, and ESCNet~\cite{ESCNet} on every applicable metric across all three test sets, and on COD10K-TE reduces $\mathcal{M}$ from 0.018 (ZoomNeXt) to 0.012, a relative improvement of $\sim$33\%.

\subsubsection{Cross-Task Architectural Transferability}

A more demanding test of the bilateral-reference paradigm is whether \emph{a single architecture}---trained separately on each benchmark, with nothing adapted per task---can match or surpass models that are purpose-built for a particular task. \cref{fig:sota_radar} visualizes this comparison, tracing $S_m$ on the $11$ primary evaluation sets of the three tasks against the strongest specialist of each.

Two observations stand out. \textbf{First}, \ourmodel{} forms the outermost contour on all $11$ axes, surpassing each task's specialist by a clear margin (average $S_m$ of $0.935$ vs $0.915$ on DIS, $0.969$ vs $0.942$ on HRSOD, and $0.929$ vs $0.897$ on COD): it does not merely transfer to other detail-demanding tasks but outperforms the task-tailored architectures on their own benchmarks. \textbf{Second}, the red contour dominates the blue contour of \ourmodelVOne{} on every axis, confirming that the gains are consistent rather than task-specific.

\subsection{Analytical Results}
\label{sec:ablation}

In this section, we analyze the contribution of each individual design choice and the sensitivity to the most important hyperparameters. Every configuration is trained on DIS5K under the final training recipe (\cref{sec:impl}) and differs from its neighbors only in the studied component. \cref{fig:ablation_visual} visualizes the component build-up with one checkpoint per configuration, on DIS-VD samples drawn from those with the largest IoU gain from the baseline framework (first row of \cref{tab:ablation_overall}) to the full model. Throughout the analytical studies (\cref{tab:ablation_overall,tab:ablation_modern,tab:ablation_ir}, the resolution study of \cref{fig:fps_vs_maxfm}, and the two dissections of the supplementary material), each reported score is the mean$\pm$std, over at least four independent training runs that share the same fixed random seed but are not bit-wise identical, of the best checkpoint of each run, with checkpoints both selected and evaluated on the DIS-VD validation set: $F_{\beta}^{x}$ is the primary readout, and every table co-reports the boundary-quality $HCE_{\gamma}$ (mean$\pm$std over the same runs). Within each run, the best checkpoint is the one with the highest $F_{\beta}^{x}$ on DIS-VD; ties at the three-decimal precision of the evaluation logs are broken by the lower $HCE_{\gamma}$ and then by the earlier epoch. \cref{tab:ablation_overall} covers the component combinations: the modernized framework enters as a single step whose derivation \cref{tab:ablation_modern} unpacks choice by choice, and all combinations of the three architectural contributions build on it---each component alone, each pair, and thereby every leave-one-out complement of the full model.
The corrections compose: the modernized framework decides \emph{what} is segmented---the baseline framework (first prediction column of \cref{fig:ablation_visual}) claims the housing row behind the boom (\emph{Truck}) and misses the camouflaged insect entirely (\emph{Locust})---the inward reference recovers filament-thin structure (the net and the thin truss bars in \emph{Hoop}, the twin boom pipes in \emph{Truck} and the leg spines in \emph{Locust}; see the insets), and the outward reference clears the spurious bucket in \emph{Ladder} (one of the few DIS-VD samples on which it changes the mask markedly), closing the gap to the ground truth.

Read along its build-up path (baseline framework $\to$ modernized framework $\to$ $+$IR $\to$ $+$IR$+$OR $\to$ full model), \cref{tab:ablation_overall} rises monotonically from 0.8722 to 0.9328 ($+$6.06 points), but its two columns tell different stories. Modernizing the framework alone lifts $F_{\beta}^{x}$ by 4.33 points while leaving boundary quality \emph{worse} than the \ourmodelVOne{} starting point ($HCE_{\gamma}$ 1270$\to$1374): a stronger encoder identifies the foreground region more reliably without recovering its fine structure. It is the inward reference that then collapses $HCE_{\gamma}$ by 45\% (1374$\to$752), at a further $+$1.30 points. Region accuracy and boundary fidelity are therefore separate axes, and a reader tracking only $F_{\beta}^{x}$ would misjudge which half of the redesign does the structural work. The remaining combination rows complete the space, and two of them locate where the detail work resides: alone on the modernized framework the dual-branch decoder is worth $+$0.0063 (0.9155$\to$0.9218), yet its marginal gain as the last step of the build-up path shrinks to $+$0.0020, so the dual-branch decoder and the inward reference overlap on the detail work; the outward reference alone reads $-$0.0010.

\begin{table*}[t]
\centering
\caption{\textbf{Component-combination ablation of the main contributions.} Every row is an independent configuration. \xmark{} denotes the \ourmodelVOne{} framework in the Modernized framework column and the \emph{absence} of the component elsewhere (no reference; a plain single-branch decoder block: two $3\times3$ convolutions, each followed by normalization, with an activation between the two, plus a $1\times1$ projection shortcut); \vonemark{} marks \ourmodelVOne{}'s counterpart of that component (\cref{tab:v1v2})}
\label{tab:ablation_overall}
\vspace{-5pt}
\renewcommand{\arraystretch}{1.0}
\setlength{\tabcolsep}{4pt}
\footnotesize
\begin{tabularx}{\textwidth}{@{}>{\columncolor{white}[0pt][\tabcolsep]}l|Y|YYY|c>{\columncolor{white}[\tabcolsep][0pt]}c@{}}
\toprule
Configuration & Modernized framework & Inward Reference (IR) & Outward Reference (OR) & Dual-Branch Decoder (DB) & $F_{\beta}^{x}\uparrow$ & $HCE_{\gamma}\downarrow$ \\
\midrule
Baseline framework & \xmark & \xmark & \xmark & \xmark & .8722$\pm$.0022 & 1270$\pm$2 \\
Modernized framework (\cref{tab:ablation_modern}) & \cmark & \xmark & \xmark & \xmark & .9155$\pm$.0024 & 1374$\pm$5 \\
\midrule
$+$ \ourmodelVOne{} references \& decoder & \cmark & \vonemark & \vonemark & \vonemark & .9193$\pm$.0028 & 1108$\pm$2 \\
\midrule
$+$ IR & \cmark & \cmark & \xmark & \xmark & .9285$\pm$.0031 & 752$\pm$7 \\
$+$ OR & \cmark & \xmark & \cmark & \xmark & .9145$\pm$.0026 & 1341$\pm$4 \\
$+$ DB & \cmark & \xmark & \xmark & \cmark & .9218$\pm$.0015 & 1297$\pm$7 \\
$+$ IR $+$ OR & \cmark & \cmark & \cmark & \xmark & .9308$\pm$.0017 & 754$\pm$6 \\
$+$ IR $+$ DB & \cmark & \cmark & \xmark & \cmark & .9300$\pm$.0024 & 731$\pm$8 \\
$+$ OR $+$ DB & \cmark & \xmark & \cmark & \cmark & .9175$\pm$.0042 & 1284$\pm$6 \\
\rowcolor{mygray}
$+$ IR $+$ OR $+$ DB (\textbf{Full model}) & \cmark & \cmark & \cmark & \cmark & \textbf{.9328$\pm$.0015} & \textbf{730$\pm$14} \\
\bottomrule
\end{tabularx}
\end{table*}

\begin{table}[t]
\centering
\caption{\textbf{From \ourmodelVOne{} to the modernized framework.} In the top block, each row replaces ($\rightarrow$) or adds ($+$) one framework choice on top of the previous row. Shaded rows mark the adopted setting of each block}
\label{tab:ablation_modern}
\vspace{-5pt}
\renewcommand{\arraystretch}{1.0}
\setlength{\tabcolsep}{3.5pt}
\footnotesize
\begin{tabularx}{\linewidth}{@{}>{\columncolor{white}[0pt][\tabcolsep]}X|c>{\columncolor{white}[\tabcolsep][0pt]}c@{}}
\toprule
Framework & $F_{\beta}^{x}\uparrow$ & $HCE_{\gamma}\downarrow$ \\
\midrule
Baseline framework & .8722$\pm$.0022 & 1270$\pm$2 \\
$\rightarrow$ DINOv3-L (frozen)        & .8870$\pm$.0018 & 1474$\pm$2 \\
$\rightarrow$ LayerNorm                & .8925$\pm$.0017 & 1454$\pm$4 \\
$\rightarrow$ GELU                     & .8948$\pm$.0021 & 1429$\pm$2 \\
$\rightarrow$ Lion \emph{(all four replacements, backbone frozen)} & .8955$\pm$.0017 & 1404$\pm$3 \\
\rowcolor{mygray}
$+$ DINOv3 last-stage tuning & .9155$\pm$.0024 & 1374$\pm$5 \\
\midrule
\multicolumn{3}{@{}l@{}}{\emph{Optimizer re-validated at the final configuration:}} \\
\rowcolor{mygray}
Full model with Lion (default)         & .9328$\pm$.0015 & 730$\pm$14 \\
Full model with AdamW                  & .9225$\pm$.0010 & 805$\pm$6 \\
\bottomrule
\end{tabularx}
\end{table}

\subsubsection{From \ourmodelVOne{} to the Modernized Framework}
\label{sec:modernize}

\cref{tab:ablation_modern} isolates the modernized framework itself. The starting point keeps every \ourmodelVOne{} framework choice---a fully fine-tuned Swin-L encoder, BatchNorm, ReLU, AdamW---but carries no bilateral references; each subsequent row then replaces one of those choices with its modern counterpart, the encoder by a frozen DINOv3-L, BatchNorm by LayerNorm, ReLU by GELU, and AdamW by Lion. The four replacements lift $F_{\beta}^{x}$ from 0.8722 to 0.8955, dominated by the encoder swap ($+$1.48); LayerNorm adds a further $+$0.55, GELU $+$0.23, and Lion $+$0.07. This configuration---a frozen encoder paired with the plain single-branch decoder---already exceeds the full \ourmodelVOne{} model on DIS-VD (0.8955 vs 0.891, cf.~\cref{tab:sota_dis}) with $\sim$$24\times$ fewer trainable parameters (8.8M) and a $12\times$ shorter schedule. We fold the four replacements, together with the last-stage tuning described below, into a single modernization step in \cref{tab:ablation_overall}.

\looseness=-1The top block of~\cref{tab:ablation_modern} closes with the DINOv3 last-stage tuning of \cref{sec:training}. Applied directly on top of the Lion row, it is the largest single step of the ladder, lifting $F_{\beta}^{x}$ from 0.8955 to 0.9155 ($+$2.00 points). Notably, it barely moves $HCE_{\gamma}$ (1404$\to$1374): the adaptation repairs the region-level fit of the frozen self-supervised features---a strong starting point, but slightly mismatched to the pixel-accurate demands of DIS---while boundary quality is delivered by the references built on top of it. Tuning depth has a single interior optimum in $F_{\beta}^{x}$, with all four of its points measured against the same full-model anchor: freezing the encoder entirely costs $-$0.0130 (0.9198$\pm$0.0028, $HCE_{\gamma}$ 807), tuning the last \emph{two} stages costs $-$0.0135 (0.9193$\pm$0.0050, $HCE_{\gamma}$ 723), and unfreezing it altogether costs $-$0.0215 (0.9113$\pm$0.0030, $HCE_{\gamma}$ 745) while also being the most expensive option. The near-symmetry of the two inner losses in $F_{\beta}^{x}$ shows that adapting exactly the last semantic stage is not a compromise between cost and accuracy but the accuracy optimum itself, and that deeper adaptation at this short schedule is both costlier and less accurate. We therefore keep this adaptation minimal and treat it as part of the modernized framework rather than a separate module.

A framework choice ultimately matters at the \emph{final} configuration, where it can disagree with the ladder---the optimizer is the case in point: with only the 8.8M-parameter decoder training, Lion merely matches AdamW, but at the full model (bottom block of \cref{tab:ablation_modern}, 84.9M trainable) Lion is decisively better---$F_{\beta}^{x}$ 0.9328$\pm$0.0015 vs 0.9225$\pm$0.0010, a gap of $1.03$ points with per-run ranges entirely disjoint (0.9310--0.9340 against 0.9210--0.9230), with $HCE_{\gamma}$ moving the same way (730 vs 805). The optimizer thus interacts with the scale of the trainable stack---inert when only the small decoder trains, decisive once the last encoder stage trains along with the complete architecture---which, together with its halved optimizer state, makes Lion the default optimizer of \ourmodel{}. The recommendation is specific to the mostly-frozen regime, and we state its converse boundary rather than leave it implicit: with the complete \ourmodel{} architecture, built with LayerNorm and GELU and mounted on a fully fine-tuned Swin-L encoder, the same optimizer diverges outright, collapsing to 0.6538$\pm$0.0066 over four runs where AdamW reaches 0.8770$\pm$0.0012.

Introduced together, the framework and the architecture raise an attribution question. The row marked ``v1'' in \cref{tab:ablation_overall} answers it directly: \ourmodelVOne{}'s own references and decoder, mounted on the same modernized framework, reach 0.9193---short of the full model by 1.35 points and 378 $HCE_{\gamma}$ counts. The three components follow \ourmodelVOne{}'s official implementation, with normalization and activation matching the host, so the gap measures architectural design under one shared recipe---not the framework choices \cref{tab:ablation_modern} already isolates. A stronger encoder and a better recipe do not, by themselves, explain \ourmodel{}---and the IR design study of \cref{tab:ablation_ir} points to the gating, not the patch layout, as the main missing piece. \ourmodelVOne{}'s components are worth just $+$0.38 in that row, and the converse transplant (reported in the supplementary material) completes the picture: our components are worth $+$0.33 on the original framework against $+$1.73 on the modernized one, and the framework itself adds $+$4.33 without our architecture against $+$5.73 with it. Both cross pairings of old and new underperform; the co-designed pair composes. Only boundary quality transfers across hosts: on the original framework the inward reference alone cuts $HCE_{\gamma}$ from 1270 to 898, and \ourmodelVOne{}'s references and decoder cut it from 1374 to 1108 on ours---as expected of mechanisms that re-inject the input image rather than refine semantics.

\begin{table}[t]
\centering
\caption{\textbf{Inward-reference design ablation.} All rows come from one dedicated sweep, so the two rows of the \emph{Inward reference} axis are independent replicates of---not identical to---the corresponding rows of \cref{tab:ablation_overall}}
\label{tab:ablation_ir}
\vspace{-5pt}
\renewcommand{\arraystretch}{1.0}
\setlength{\tabcolsep}{3pt}
\footnotesize
\begin{tabularx}{\linewidth}{@{}>{\columncolor{white}[0pt][\tabcolsep]}lX|c>{\columncolor{white}[\tabcolsep][0pt]}c@{}}
\toprule
Design axis & Value & $F_{\beta}^{x}\uparrow$ & $HCE_{\gamma}\downarrow$ \\
\midrule
                                     & absent                        & .9156$\pm$.0022 & 1374$\pm$6 \\
\rowcolor{mygray}
\cellcolor{white}\multirow{-2}{*}{Inward reference} & adopted            & .9332$\pm$.0040 & 751$\pm$10 \\
\midrule
\rowcolor{mygray}
\cellcolor{white}                    & hybrid Window/Conv            & .9332$\pm$.0040 & 751$\pm$10 \\
                                     & Conv SGG only                 & .9266$\pm$.0029 & 757$\pm$5 \\
\multirow{-3}{*}{Gate (SGG)}         & none (ungated)                & .9260$\pm$.0024 & 993$\pm$6 \\
\midrule
                                     & grid split ($\alpha{=}0$)     & .9300$\pm$.0007 & 814$\pm$7 \\
                                     & space-to-depth ($\alpha{=}0$) & .9298$\pm$.0029 & 750$\pm$5 \\
\rowcolor{mygray}
\cellcolor{white}\multirow{-3}{*}{Layout} & space-to-depth ($\alpha{=}0.2$) & .9332$\pm$.0040 & 751$\pm$10 \\
\midrule
                                     & 0                             & .9298$\pm$.0029 & 750$\pm$5 \\
                                     & 0.1                           & .9318$\pm$.0022 & 745$\pm$7 \\
\rowcolor{mygray}
\cellcolor{white}                    & 0.2                           & .9332$\pm$.0040 & 751$\pm$10 \\
\multirow{-4}{*}{Overlap ratio $\alpha$} & 0.4                       & .9288$\pm$.0028 & 751$\pm$8 \\
\midrule
                                     & 16                            & .9292$\pm$.0011 & 753$\pm$6 \\
\rowcolor{mygray}
\cellcolor{white}                    & 64                            & .9332$\pm$.0040 & 751$\pm$10 \\
                                     & 256                           & .9300$\pm$.0029 & 747$\pm$4 \\
\multirow{-4}{*}{Compression $d_c$}  & off (uncompressed)            & .9280$\pm$.0021 & 755$\pm$9 \\
\midrule
                                     & 64                            & .9298$\pm$.0018 & 766$\pm$8 \\
\rowcolor{mygray}
\cellcolor{white}                    & 96                            & .9332$\pm$.0040 & 751$\pm$10 \\
                                     & 128                           & .9308$\pm$.0013 & 740$\pm$3 \\
\multirow{-4}{*}{Detail channels $C_r$} & pyramid ($0.2\times$ neck) & .9270$\pm$.0020 & 780$\pm$2 \\
\bottomrule
\end{tabularx}
\end{table}

\subsubsection{Effectiveness of Inward Reference}

Activating the redesigned inward reference on top of the modernized framework (\cref{tab:ablation_overall}) lifts $F_{\beta}^{x}$ from 0.9155 to 0.9285---the largest architectural increment ($+$1.30 points)---while cutting $HCE_{\gamma}$ from 1374 to 752, nearly halving the residual correction effort. The leave-one-out view (the ``$+$ OR $+$ DB'' row of \cref{tab:ablation_overall}) confirms the same picture from the opposite direction, and more sharply: removing the inward reference from the full model is the costliest single removal on both readouts, $-$0.0153 on $F_{\beta}^{x}$ and $HCE_{\gamma}$ from 730 to 1284. Removing only the gate while keeping the detail stream (the ungated $\mathbf{R}_s^{\text{in}}$) costs $-$0.0043 on $F_{\beta}^{x}$ (0.9285$\pm$0.0039) and degrades $HCE_{\gamma}$ to 918: the gate moves both readouts, and its dominant effect is on \emph{how cleanly} the structure is reconstructed---precisely the quantity the DIS5K protocol measures with $HCE_{\gamma}$---and together the detail stream and its gate act as the model's boundary-quality engine. A further control dissects the hybrid gating that \cref{fig:sgg} depicts: removing only the full-resolution stage-$s_4$ injection site costs $-$0.0045 on $F_{\beta}^{x}$ but inflates $HCE_{\gamma}$ from 730 to 1096 ($+$50\%)---the native-resolution entry point itself, not only its gate, carries the~boundary-quality~gain.

\cref{tab:ablation_ir} dissects the module's internal design where the inward reference enters the chain---the modernized framework in place, the outward reference and the dual-branch decoder absent---deviating from the adopted configuration one axis at a time. The sweep independently reproduces the module-level step ($+$1.76 points, $HCE_{\gamma}$ 1374$\to$751) and shows which of the internal choices carry weight. The gate is the sharpest axis. Removing it costs $-$0.0072 and inflates $HCE_{\gamma}$ by 242 counts---over one and a half times what the same removal costs at the full-model operating point ($-$0.0043)---so the gate matters most at the point where the detail stream first enters. Replacing the Window SGG of stages $s_0$--$s_3$ by the convolutional form costs $-$0.0066 at essentially unchanged boundary quality (757 vs 751): the window form buys region-level accuracy that the convolutional one does not, while stage $s_4$ keeps the convolutional form because window attention at full resolution costs about a third more peak training memory per sample (\cref{sec:inward}). The three swept axes are each single-peaked at the adopted value---overlap $\alpha{=}0.2$ ($-$0.0034 at $0$, $-$0.0044 at $0.4$), compression $d_c{=}64$, and detail budget $C_r{=}96$---and two of them carry a lesson beyond tuning: leaving the patches uncompressed, the widest and most expensive option, is the worst point of its axis ($-$0.0052), and a pyramid-shaped detail budget mirroring the neck is worse than a uniform one on both readouts ($-$0.0062, $HCE_{\gamma}$ 780). At matched zero overlap the space-to-depth layout ties \ourmodelVOne{}'s grid split on $F_{\beta}^{x}$ (0.9298 vs 0.9300) but is 64 counts better in $HCE_{\gamma}$, and only space-to-depth admits overlap at all: the locality-preserving layout pays off in boundary quality and in the design freedom it unlocks. Boundary quality is in fact governed by a different subset of the choices than region accuracy---$HCE_{\gamma}$ varies by ten counts across the overlap and compression axes (745--755) but responds strongly to the presence of the gate, to the layout, and to how the detail budget is distributed.

\begin{table}[t!]
\centering
\caption{\textbf{Scaling the semantic encoder on DIS5K.} Each size reports its best checkpoint, selected by the five-set mean of $F_{\beta}^{x}$. Params: total (trainable), in M; TMACs at $1024^2$}
\label{tab:scaling}
\vspace{-10pt}
\renewcommand{\arraystretch}{1.0}
\setlength{\tabcolsep}{3pt}
\footnotesize
\begin{tabularx}{\linewidth}{@{}>{\columncolor{white}[0pt][\tabcolsep]}l|Z{1.2}Z{0.8}Z{0.6}|Z{1.2}>{\columncolor{white}[\tabcolsep][0pt]\hsize=1.2\hsize\centering\arraybackslash}X@{}}
\toprule
 &  &  &  & \multicolumn{2}{c}{$F_{\beta}^{x}\uparrow\,/\,HCE_{\gamma}\downarrow$} \\
\cmidrule(l){5-6}
DINOv3 & Params & TMACs & FPS & DIS-TEs & DIS-VD \\
\midrule
S+          & 35 (16)     & 0.39  & 34.1 & .911\,/\,792 & .907\,/\,818 \\
B           & 94 (37)     & 0.79  & 29.8 & .917\,/\,763 & .916\,/\,756 \\
\rowcolor{mygray}
L (default) & 312 (85)    & 2.21  & 18.4 & .936\,/\,736 & .932\,/\,731 \\
H+          & 851 (220)   & 4.97  & 11.9 & .938\,/\,729 & .930\,/\,745 \\
7B          & 6738 (1700) & 33.27 & 2.5  & .939\,/\,733 & .942\,/\,739 \\
\bottomrule
\end{tabularx}
\end{table}

\subsubsection{Effectiveness of Outward Reference}

The outward reference's measured effect on the region metric is small at every operating point of \cref{tab:ablation_overall}, and we report it as such: enabling it on top of the modernized framework$+$IR stack lifts $F_{\beta}^{x}$ from 0.9285 to 0.9308 while $HCE_{\gamma}$ rises from 752 to 754, removing it from the full model (the ``$+$ IR $+$ DB'' row) costs $-$0.0028 and leaves $HCE_{\gamma}$ one count \emph{higher}, and alone on the modernized framework it reads $-$0.0010 (0.9145$\pm$0.0026). We keep it because it completes the paradigm this paper re-instantiates: the outward direction supplies boundary evidence that the mask loss alone does not carry, and the supplementary material shows the same components behave very differently depending on what they are attached to. The learned boundary tracks its prediction-gated target closely across DIS-VD (Pearson $r{=}0.67{\pm}0.05$ at $s_2$), and the resulting residual attention concentrates on the fine structures it serves.

\subsubsection{Effectiveness of Dual-Branch Decoder}

Replacing the plain single-branch decoder block with our dual-branch decoder block completes the full model: $F_{\beta}^{x}$ rises from 0.9308 to 0.9328 with the lowest $HCE_{\gamma}$ of \cref{tab:ablation_overall} (730). The combination rows of \cref{tab:ablation_overall} locate its contribution precisely: alone on the modernized framework the block is worth $+$0.0063 (0.9155$\to$0.9218, with $HCE_{\gamma}$ 1374$\to$1297)---the largest single-component gain after the inward reference---although its marginal gain as the last step of the build-up path shrinks to $+$0.0020---direct evidence that part of its contribution overlaps with the inward reference's detail stream. We keep the block for its cumulative gain and its role as the structural link between the two references. Retaining only the boundary branch costs 0.0033 on the region metric (0.9295) and only the context branch 0.0015 (0.9313), against 0.9328 for the pair, and $HCE_{\gamma}$ orders them the same way (772 and 741 against 730): the pair is the cleanest configuration. Replacing the large-kernel attention with an $O(N^2)$ boundary-aware self-attention context branch reads $-$0.0020 on $F_{\beta}^{x}$ and lowers $HCE_{\gamma}$ from 730 to 717, at a substantially higher compute cost, supporting the~linear-cost~choice.

\subsubsection{Scaling the Semantic Encoder}
\label{sec:scaling}

Because everything except the encoder is held fixed---the decoder, both references, the last-stage tuning rule, and the final training recipe---swapping the mostly-frozen DINOv3 backbone yields a controlled, and unusually affordable, scaling study for high-resolution segmentation (\cref{tab:scaling}). Three observations emerge. \emph{First}, semantic capacity pays steeply up to the L scale, then returns diminish sharply: on the four DIS-TE splits the readout climbs from 0.911 (S+) through 0.917 (B) to 0.936 at L---$+$2.5 points across a $9\times$ span of model size---after which the $2.7\times$ larger H+ adds $+$0.20 points and the $22\times$ larger 7B $+$0.33 points. On DIS-VD the top of the family reads $-$0.2 points for H+ and $+$1.0 points for 7B. Boundary quality traces the same shape (DIS-TEs $HCE_{\gamma}$ 792$\to$763$\to$736 up to L, then 729 and 733). \emph{Second}, mechanism beats raw capacity: past the default L, a $2.7\times$--$22\times$ larger model buys at most $+$0.33 points on the DIS-TE splits and $+$1.0 points on DIS-VD, whereas our architecture contributes $+$1.7 points on DIS-VD at fixed scale (\cref{tab:ablation_overall})---so L remains the accuracy/cost sweet spot and our default, within seven $HCE_{\gamma}$ counts of the family's best. \emph{Third}, the family spans practical deployment endpoints: the 35M S+ variant reaches 0.907 on DIS-VD at 34.1\,FPS---above the fully fine-tuned Swin-L \ourmodelVOne{} (0.891, \cref{tab:sota_dis})---while the 6.7B variant still runs single-forward at 2.5\,FPS with a 14.5\,GB peak, roughly $8\times$ faster and slightly leaner than DiffDIS's official inference (\cref{tab:efficiency}). Encoder scale thus forms a second deployment axis, orthogonal to the resolution axis studied next: resolution sets the spatial token budget, and encoder size the per-token semantic capacity.

\begin{table}[t!]
\centering
\caption{\textbf{Efficiency comparison.} Each method runs its official inference path ($1024\times1024$ input, batch size 1, bf16 mixed precision); the Trainable column counts the inference model's parameters that are trained}
\label{tab:efficiency}
\vspace{-5pt}
\renewcommand{\arraystretch}{1.0}
\setlength{\tabcolsep}{2.5pt}
\footnotesize
\begin{tabularx}{\linewidth}{@{}>{\columncolor{white}[0pt][\tabcolsep]}Xrrr>{\columncolor{white}[\tabcolsep][0pt]}r@{}}
\toprule
Method & \#Params (M)$\downarrow$ & Trainable$\downarrow$ & FPS$\uparrow$ & Peak mem.\ (GB)$\downarrow$ \\
\midrule
IS-Net$_{22}$~\cite{DIS5K}            & 44.0 & 44.0 & 77.4 & 2.8 \\
InSPyReNet$_{22}$~\cite{InSPyReNet}   & 90.7 & 90.7 & 24.9 & 2.8 \\
UDUN$_{23}$~\cite{UDUN}               & 24.9 & 24.9 & 63.0 & 0.6 \\
MVANet$_{24}$~\cite{MVANet}           & 94.1 & 94.1 & 19.1 & 2.6 \\
GenPercept$_{25}$~\cite{GenPercept}   & 949.6$^\dagger$ & 865.9 & 3.9 & 8.8 \\
DiffDIS$_{25}$~\cite{DiffDIS}         & 996.6$^\dagger$ & 912.9 & 0.31$^\ddagger$ & 16.7 \\
PDFNet$_{26}$~\cite{PDFNet}           & 433.2 & 97.9 & 1.9$^\P$ & 9.5 \\
\midrule\midrule
ZoomNeXt$_{24}$~\cite{ZoomNeXt}       & 84.8 & 84.8 & 11.7 & 1.4 \\
FSEL$_{24}$~\cite{FSEL}               & 67.3 & 67.3 & 20.3 & 1.1 \\
ESCNet$_{25}$~\cite{ESCNet}           & 99.9 & 99.9 & 21.0 & 1.6 \\
RUN$_{25}$~\cite{RUN}                 & 79.2 & 79.2 & 6.9 & 6.1 \\
\midrule
\ourmodelVOne{}$_{24}$~\cite{BiRefNet}       & 215   & 215   & 15.3 & 3.5 \\
\rowcolor{mygray}
\textbf{\ourmodel{}}                  & 312.4 & 84.9 & 18.4 & 3.2 \\
\bottomrule
\multicolumn{5}{@{}p{0.98\linewidth}@{}}{\rule{0pt}{1.05em}\scriptsize $^\dagger$Excluding the frozen text encoder ($\sim$340\,M). $^\ddagger$The official inference includes 6-pass test-time augmentation; a single pass runs at 1.9\,FPS. $^\P$The official two-stage inference includes the same 6-pass test-time augmentation and pseudo-depth generation by a 335\,M Depth Anything V2 estimator, matching the accounting reported in its paper; a single pass runs at 9.7\,FPS.}\\
\end{tabularx}
\end{table}

\subsubsection{Efficiency}
\label{sec:efficiency}

\textbf{Model size and GMACs.}~\ourmodel{} contains 312.4M total parameters, of which 84.9M are trainable (9.3M in the decoder plus the fine-tuned last backbone stage; the remaining 227.5M are frozen). At the standard input resolution of $1024\times1024$, the model performs 2208.8 GMACs ($\approx$2.2\,TMACs) per forward pass, of which 2068.3 GMACs reside in the mostly-frozen DINOv3-L backbone and only 140.5 GMACs in the decoder stack. Compared with the previous \ourmodelVOne{}, which trains the entire encoder--decoder stack (215M trainable parameters~\cite{BiRefNet}), \ourmodel{} cuts the trainable parameter count to $\sim$40\% while delivering the accuracy gains documented in~\cref{tab:sota_dis,tab:sota_hrsod,tab:sota_cod}. \ourmodel{} thus carries more total parameters than \ourmodelVOne{} but optimizes far fewer, making training substantially cheaper: a $12\times$ shorter schedule and a modest per-GPU memory footprint (detailed below). Tighter compute budgets are served by the backbone family of \cref{tab:scaling} (down to 0.39\,TMACs at S+) and by the synchronized low-resolution regime below.

\textbf{Inference efficiency.}~\ourmodel{} runs at 18.4\,FPS with bf16 mixed precision and batch size 1 at $1024\times1024$ input, with a peak GPU memory footprint of $\sim$3.2\,GB. \cref{tab:efficiency} places this figure in context: every compared method listed there is re-benchmarked under the single protocol of the table caption, keeping each method's official inference procedure intact (\eg, the test-time augmentation built into the DiffDIS release). \ourmodel{} runs faster and leaner than \ourmodelVOne{} (18.4 vs 15.3\,FPS; 3.2 vs 3.5\,GB) despite carrying a larger total parameter count; lightweight CNN-based detectors such as IS-Net remain $\sim$$4\times$ faster, but at a substantial accuracy deficit (\cref{tab:sota_dis}). PDFNet, the strongest competitor on the primary DIS metric, pays for its pseudo-depth prior with a 433\,M two-stage pipeline running at 1.9\,FPS---an order of magnitude slower than \ourmodel{}. The contrast with the diffusion-based family is starkest: GenPercept needs $\sim$950\,M inference parameters for 3.9\,FPS, and DiffDIS---the strongest fully automatic diffusion-based method in our comparison---runs its official inference at 0.31\,FPS with a 16.7\,GB peak, roughly $60\times$ slower and $5\times$ more memory-hungry than \ourmodel{} while trailing it in accuracy (\cref{tab:sota_dis}). Single-forward discriminative inference thus remains the only route that combines top accuracy with a commodity-hardware footprint at full resolution. On the COD side the pattern repeats: the strongest specialist, ZoomNeXt, runs at 11.7\,FPS and the Mamba-based RUN at 6.9\,FPS, while \ourmodel{} serves COD with the identical architecture and cost.

\begin{figure}[t!]
\centering
\includegraphics[width=.8\linewidth]{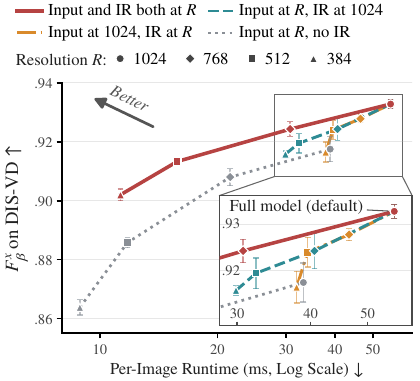}
\vspace{-5pt}
\caption{\textbf{Resolution--accuracy trade-off on DIS-VD.} Each point is a model \emph{retrained from scratch} at its resolution configuration. The inset's runtime axis is also logarithmic. Error bars: $\pm$1 standard deviation.}
\label{fig:fps_vs_maxfm}
\end{figure}

\begin{figure*}[t!]
\centering
\includegraphics[width=\linewidth]{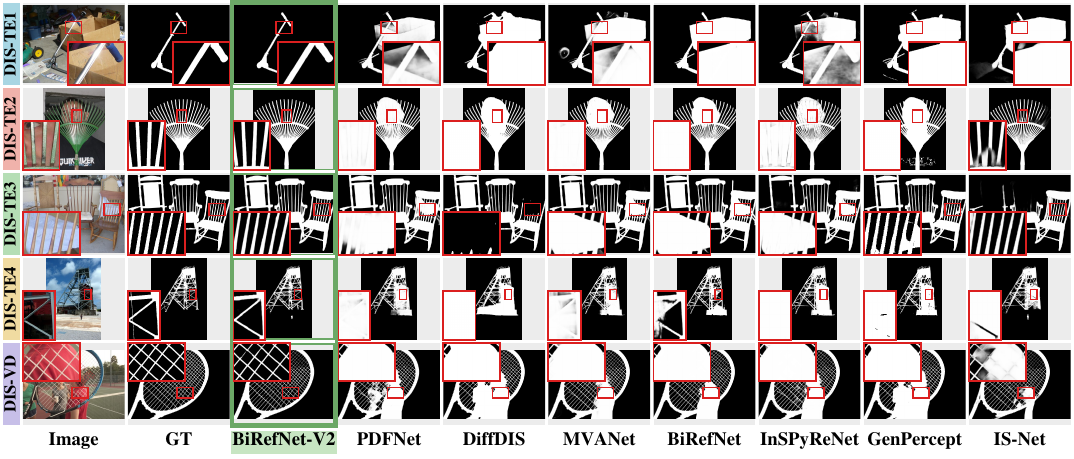}
\vspace{-15pt}
\caption{\textbf{Qualitative comparison on DIS5K.} An all-white inset means the method labels the whole enlarged region as foreground.}
\label{fig:qual_dis}
\end{figure*}

\textbf{Resolution--accuracy trade-off.}~Because the resolutions of the input (backbone) stream and the inward-reference detail stream can be set independently, deployment raises two questions: how accuracy trades against throughput as the input scales down, and whether the two streams should be scaled together or independently. We answer both by \emph{retraining} the final architecture from scratch at every resolution configuration and plotting DIS-VD $F_{\beta}^{x}$ against per-image runtime (\cref{fig:fps_vs_maxfm}). Three findings emerge. \emph{First}, synchronized down-scaling is a well-behaved deployment frontier: $768^2$ runs in $31$\,ms/image ($1.8\times$ faster than the full model) within $0.9$ points of it ($0.924$ vs $0.933$), $512^2$ in $16$\,ms at $-2.0$ points, and $384^2$ in $11$\,ms at $-3.1$ points. \emph{Second}, scaling both streams together is what turns resolution into speed: decoupling---shrinking only the input while the inward reference stays at $1024^2$, or the reverse---keeps accuracy within $1.7$ points of the full model even at $384^2$, but cannot bring per-image runtime below $\sim$$30$--$47$\,ms because the stream left at $1024^2$ remains the compute floor. Between the two decoupled regimes, however, there is no consistent preference to state: at matched resolution $R$ the reverse direction leads by 0.0007--0.0045, but that ordering inverts once the two are compared at matched \emph{runtime}---the fair comparison, since down-scaling the heavy semantic stream is far cheaper than down-scaling the light detail stream---where shrinking the input instead leads by up to 0.0063 at the fastest budget the two regimes share. Decoupling is therefore a way to preserve accuracy, not a way to buy throughput, and which stream to keep at full resolution can be decided on engineering grounds rather than on accuracy. \emph{Third}, the injected detail earns its cost throughout: the no-IR floor (the inward reference removed) trails the synced curve at every operating point, and the gap widens sharply as the backbone loses resolution---$1.6$ points at $768^2$, $2.8$ at $512^2$, $3.8$ at $384^2$---so the detail stream acts as a resolution-adaptive injector that matters most where aggressive downsampling discards the most high-frequency content.
At $1024^2$ no additional training is needed: the anchor shared by the three regimes with the inward reference is the full model and the endpoint of the no-IR curve is the ``$+$ OR $+$ DB'' configuration, so both points reuse the from-scratch runs behind these two rows of \cref{tab:ablation_overall}. The runtimes in \cref{fig:fps_vs_maxfm} were measured earlier than those of \cref{tab:efficiency}, with the same precision and batch size but on an older software stack with a shorter timing loop, so the full model's runtime there corresponds to 18.0 rather than its tabulated 18.4\,FPS.

\textbf{Training efficiency.}~The mostly-frozen-backbone design also shrinks the training cost: \ourmodel{} converges in 50 epochs on DIS5K (versus 600 epochs for \ourmodelVOne{}, a $12\times$ reduction), with a per-GPU footprint of 6.5\,GB at batch size 1 and 11.4\,GB at batch size 2 ($1024\times1024$, bf16)---within consumer-grade 32\,GB cards. We attribute this efficiency to two factors: (i) gradients flow only through the decoder and the last backbone stage, removing most of the backbone's share of the activation memory; and (ii) the strong DINOv3 features remove the need for the long schedules that fully fine-tuned encoders need to become foreground-aware.

\textbf{Maximum native resolution.}~The same mostly-frozen-backbone economy also raises the ceiling for \emph{native} high-resolution training: the 32\,GB card used above sustains full training steps at up to $2432\times2432$ (forward-only inference up to $3968\times3968$), and a large-memory 268\,GB GPU trains at up to $7552\times7552$ (see the supplementary material).

\subsection{Qualitative Results}
\label{sec:qualitative}

\begin{figure*}[t]
\centering
\includegraphics[width=\linewidth]{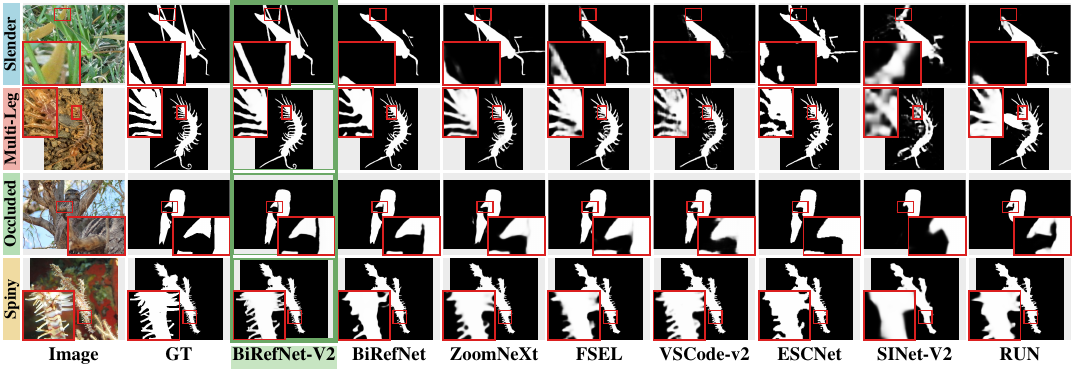}
\vspace{-15pt}
\caption{\textbf{Qualitative comparison on COD.} All four samples are from COD10K-TE.}
\label{fig:qual_cod}
\end{figure*}

To complement the quantitative numbers, \cref{fig:qual_dis,fig:qual_cod} compare \ourmodel{} against representative state-of-the-art methods on DIS5K and COD, with our prediction placed next to the ground truth in every panel.

\noindent\textbf{DIS5K (\cref{fig:qual_dis}).}
The five rows span the benchmark's full difficulty spectrum, from DIS-TE1 (top) to DIS-VD (bottom), and separate two failure modes. The first is semantic: the annotated foreground of the DIS-TE1 image is the kick scooter alone, whereas \emph{every} competitor also claims the cardboard box behind it. The second is structural and recurs on rake tines (TE2), chair spindles (TE3), a steel tower's cross-bracing (TE4), and a racket's string bed (VD): inside the red box the openwork is filled into a solid patch, washed out, or erased outright by most competitors, the strongest included, while \ourmodel{} keeps every gap open. The red insets enlarge the highest-frequency or most semantically confusable sub-regions; the component-level evidence for the locality-preserving inward reference and its gate is given by \cref{tab:ablation_overall,tab:ablation_ir} and \cref{fig:ablation_visual}.

\noindent\textbf{COD (\cref{fig:qual_cod}).}
The four rows isolate four distinct fine-structure challenges. \ourmodel{} traces the grasshopper's slender hind legs all the way to the image border, which every competing method truncates or drops (\emph{Slender}); resolves the centipede's legs individually, which \emph{RUN} partly fuses with an adjacent gray object and \emph{SINet-V2} smears into a ghost (\emph{Multi-Leg}); reproduces both the branch crossing the owl's body and the tail tapering below it, which \emph{SINet-V2} blurs away and \emph{RUN} drops altogether (\emph{Occluded}); and keeps the comb of fine spines along the ghost pipefish's underside, which \emph{SINet-V2} smooths into a featureless outline (\emph{Spiny}). Across both benchmarks, the qualitative gap is thus widest exactly on the structures that the redesigned inward reference, the outward reference, and the dual-branch decoder jointly target---filament-thin, densely articulated, occluded, or finely spined foregrounds---explaining why the quantitative gains (\cref{tab:sota_dis,tab:sota_cod}) translate into visibly more~reliable~masks.

\section{Applications}
\label{sec:applications}

\begin{figure}[t!]
\centering
\includegraphics[width=\linewidth]{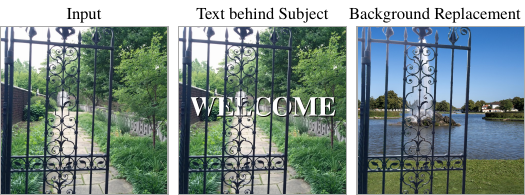}
\caption{\textbf{Application demonstration.} A held-out DIS-TE4 wrought-iron gate processed by the released \ourmodel{}. New background: cropped from a photo by \href{https://commons.wikimedia.org/?curid=186918966}{AuHaidhausen}, \href{https://creativecommons.org/licenses/by/4.0/}{CC~BY~4.0}.}
\label{fig:applications}
\end{figure}

High-resolution dichotomous segmentation has moved well beyond the laboratory. Since its original release in 2024, the open-source \ourmodelVOne{} family has accumulated more than 27 million downloads on Hugging Face\footnote{\url{https://huggingface.co/api/models/ZhengPeng7/BiRefNet?expand[]=downloadsAllTime}: the Hugging Face API reports all-time downloads per model---20.1M for this flagship repository alone, 27.0M summed across the official model series as of October 2026.} and has seen third-party adoption and recognition: Cloudflare adopted the model to power the beta background-removal feature of its Images API\footnote{\url{https://blog.cloudflare.com/background-removal} (accessed October 2026): ``For our beta version, we opted to use the BiRefNet model to drive decisions for background removal''; ``The Images API now supports automatic background removal for hosted and remote images.''}, and OpenCV includes the model in the official benchmark suite of its new OpenCV~5 DNN engine\footnote{\url{https://github.com/opencv/opencv/wiki/OpenCV-5-DNN-Benchmarks} (accessed October 2026): \ourmodelVOne{} is among the models profiled in the official OpenCV~5 DNN benchmarks.}. Because \ourmodel{} preserves its predecessor's single-forward interface (\cref{sec:inference,sec:efficiency}), existing integrations can adopt it without interface changes and inherit the accuracy and efficiency gains reported in \cref{tab:sota_dis,tab:sota_hrsod,tab:sota_cod} and in \cref{sec:efficiency}.

For deployment, three properties of \ourmodel{} matter beyond benchmark scores. First, adaptation is light: a complete training run needs 50 epochs ($\sim$8\% of its predecessor's schedule), and training fits in 6.5\,GB of GPU memory at batch size 1 and $1024\times1024$ (\cref{sec:efficiency}), so fine-tuning on data of comparable scale costs at most one such run---feasible on commodity hardware. Second, the same memory economy puts the native-resolution ceiling at $2432\times2432$ for training and $3968\times3968$ for inference on a single 32\,GB consumer-grade GPU (\cref{sec:efficiency}). Third, the operating point is tunable along two orthogonal axes: synchronized resolution scaling converts resolution into throughput ($1.8\times$ faster at $768\times768$ within 0.9 points, \cref{fig:fps_vs_maxfm}), while the encoder family of \cref{tab:scaling} trades compute for semantic capacity, spanning 35M parameters (34.1\,FPS at $1024\times1024$) to 6.7B under one decoder---lower the resolution when latency dominates, and scale the encoder when accuracy is the priority.

\cref{fig:applications} shows two canonical edits driven by one predicted mask: text insertion behind the subject and full background replacement---both hinge on keeping intact the openwork structures that motivated this work. In the background replacement, the composite keeps the input's own pixel colors, so the wrought-iron gate's upper-left bars and finials, photographed against a bright sky, stay pale over the new background. We expect the lowered training and deployment cost to widen the model's adoption, from interactive editors to automated content pipelines.

\begin{figure}[t!]
\centering
\includegraphics[width=\linewidth]{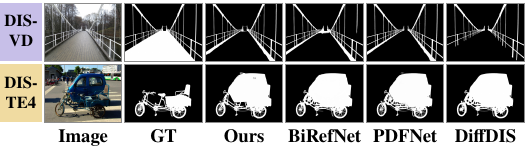}
\vspace{-20pt}
\caption{\textbf{Failure cases on DIS5K.} Top: ambiguous extent of the target; bottom: same-color adjacency.}
\label{fig:failure}
\end{figure}

\section{Limitations}

\cref{fig:failure} shows two characteristic failure cases, drawn from the worst predictions of \ourmodel{} on DIS5K as ranked by per-image $\mathcal{M}$. Both errors lie at the level of foreground \emph{definition}---deciding what counts as the object---rather than segmentation quality: the model truncates a target whose extent is genuinely ambiguous (the bridge deck), and merges a second, same-colored vehicle into the target (the cargo tricycle). \ourmodelVOne{}, PDFNet, and DiffDIS, shown alongside, fail in the same way in both cases: the errors are neither introduced by our redesign nor specific to the bilateral-reference paradigm---they arise where the foreground decision itself is the hard part. The pattern is consistent with the design: the references chiefly sharpen \emph{how well} the selected region is delineated and only occasionally prune a spurious region (the bucket in \emph{Ladder}, \cref{fig:ablation_visual}), while the selection itself is made by the semantic stream, whose verdict the Semantically Guided Gate can only execute, not overturn; when that stream commits to the wrong target, detail injection refines the wrong decision. The semantic axis, on which \ourmodel{} also improves over prior methods (\cref{fig:qual_dis}, top row), is thus where its residual failures concentrate: the paradigm moves the frontier from delineation to definition.

Beyond these failure modes, \ourmodel{} has two systemic limitations. First, it inherits the inference cost of its ViT-L backbone: although mostly frozen at training time, the backbone accounts for 2068.3 of the 2208.8 GMACs per forward pass, limiting full-resolution inference to 18.4\,FPS; latency-critical deployments must fall back to the synced low-resolution regime of \cref{sec:efficiency} or to distillation. Second, training remains at $1024\times1024$, which under-exploits natively 4K benchmarks such as UHRSD-TE (\cref{tab:sota_hrsod}).

\section{Conclusion}
We have presented \ourmodel{}, which recasts the bilateral-reference paradigm for high-resolution DIS as \emph{controlled information recovery}: raw image detail re-enters the decoder locally, selectively, and under supervision focused on the object being segmented. Tracing the shortfalls of the original \ourmodelVOne{} to these three unmet requirements, we redesigned both references and the decoder block that links them, and paired them with a modernized, mostly-frozen framework. The resulting model leads across all $13$ evaluation sets of three tasks and converges in an order of magnitude fewer epochs than its predecessor.

The broader lesson is a separation of concerns: a heavy semantic stream that can stay mostly frozen, and a light detail stream that runs at full resolution under semantic gating. This separation lets the framework absorb stronger frozen encoders (\cref{sec:scaling}) and set each stream's resolution independently (\cref{fig:fps_vs_maxfm}). We see the paradigm as a reusable template for dense prediction tasks whose difficulty lies in fine structure rather than semantics, with alpha matting and medical segmentation as immediate candidates.

{
\bibliographystyle{IEEEtran}
\bibliography{main}
}

\vspace{-8mm}
\begin{IEEEbiography}[{\includegraphics[width=1in,height=1.25in,clip,keepaspectratio]{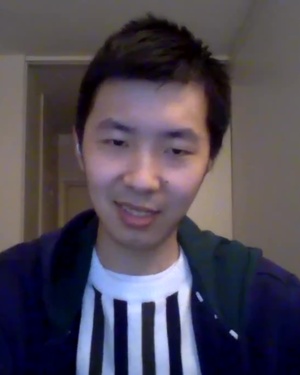}}]{Peng Zheng}
received the M.Sc. degree in ICT Innovation (Visual Computing and Communication) from Aalto University, Finland, and the University of Trento, Italy, in 2023. He is currently working toward the PhD degree with the Multimedia and Human Understanding Group (MHUG), University of Trento, Italy. He was a research intern with the Inception Institute of Artificial Intelligence (IIAI) from 2021 to 2022, a research assistant with MBZUAI in 2022, and a research intern with Alibaba Group in 2023. He is the author of the open-source BiRefNet series for high-resolution segmentation, with more than 20 million downloads on Hugging Face. He serves as a reviewer for journals and conferences including IEEE TPAMI, IJCV, TIP, CVPR, ICCV, ECCV, and NeurIPS. His research interests include computer vision and machine learning.
\end{IEEEbiography}

\vspace{-8mm}
\begin{IEEEbiography}[{\includegraphics[width=1in,height=1.25in,clip,keepaspectratio]{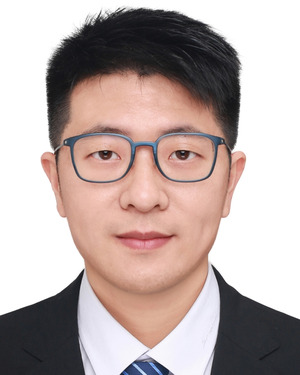}}]{Deng-Ping Fan}
received the PhD degree from Nankai University in 2019, under the supervision of Prof. Ming-Ming Cheng. He is a full professor and the deputy director with the Media Computing Lab (MCLab), College of Computer Science, Nankai University, China. Prior to his current position, he was a postdoctoral fellow, collaborating with Prof. Luc Van Gool in the Computer Vision Lab, ETH Zurich. From 2019 to 2021, he served as a research scientist (Principal Investigator) and team leader of IIAI - CV\&Med at the Inception Institute of Artificial Intelligence, working alongside Prof. Ling Shao and Prof. Jianbing Shen. His research interests span computer vision, machine learning, and medical image analysis. Specifically, he focuses on Dichotomous Image Segmentation and Multi-Modal AI. He is a senior member of IEEE.
\end{IEEEbiography}

\vspace{-8mm}
\begin{IEEEbiography}[{\includegraphics[width=1in,height=1.25in,clip,keepaspectratio]{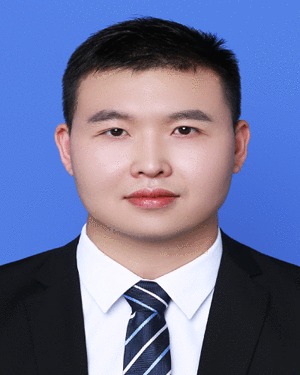}}]{Binglu Wang}
(M'21) received the Ph.D. degree in Control Science and Engineering from the School of Automation at Northwestern Polytechnical University, Xi'an, China, in 2021. He was a postdoctoral researcher with the School of Information and Electronics, Beijing Institute of Technology, Beijing, China, from 2022 to 2024. He is currently a professor with the School of Astronautics, Northwestern Polytechnical University. His research interests include computer vision, digital signal processing, and deep learning.
\end{IEEEbiography}

\vspace{-8mm}
\begin{IEEEbiography}[{\includegraphics[width=1in,height=1.25in,clip,keepaspectratio]{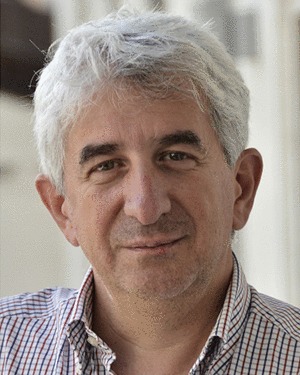}}]{Nicu Sebe}
is a Professor at the University of Trento, Italy, leading the research in the areas of multimedia information retrieval and human behavior understanding as the leader of the Multimedia and Human Understanding Group (MHUG). He was a Program Chair of the ACM Multimedia 2011 and 2007, ECCV 2016, ICCV 2017, and ICPR 2020. He is a Program Chair of CVPR 2027 and a General Chair of ACM MM 2027 and ECCV 2028. He is the co-editor in chief of Computer Vision and Image Understanding. He is a senior member of IEEE and ACM and a fellow of IAPR and ELLIS.
\end{IEEEbiography}

\vfill

\end{document}

%% file: utils/preamble.tex
\usepackage{anyfontsize}
\usepackage{amsmath,amsfonts,amssymb}
\usepackage{algorithmic}
\usepackage{algorithm}
\usepackage{array}
\usepackage{tabularx}
\usepackage[caption=false,font=normalsize,labelfont=sf,textfont=sf]{subfig}
\usepackage{textcomp}
\usepackage{stfloats}
\fnbelowfloat
\usepackage{url}
\usepackage{ragged2e}
\usepackage{verbatim}
\usepackage{graphicx}
\usepackage[nocompress]{cite}
\usepackage{multirow}
\usepackage[table,dvipsnames]{xcolor}
\usepackage{makecell}
\usepackage{booktabs}
\usepackage{overpic}

\def\ie{\emph{i.e.}}
\def\eg{\emph{e.g.}}

\def\etal{{\em et al.}}

\definecolor{mygray}{gray}{.92}
\definecolor{myRed}{RGB}{219, 68, 55}
\definecolor{myGreen}{RGB}{15, 157, 88}
\definecolor{myBlue}{RGB}{66, 133, 244}

\def\ourmodel{BiRefNet-V2}
\newcommand{\ourmodelVOne}{BiRefNet}

\newcommand{\cmark}{\pmb{\checkmark}}
\newcommand{\xmark}{\textcolor{black!25}{\footnotesize$\times$}}
\newcommand{\vonemark}{\textcolor{black!50}{\footnotesize v1}}
\newcolumntype{Y}{>{\centering\arraybackslash}X}
\newcolumntype{Z}[1]{>{\hsize=#1\hsize\centering\arraybackslash}X}

\makeatletter
\newcommand{\thickhline}{%
    \noalign {\ifnum 0=`}\fi \hrule height 1pt
    \futurelet \reserved@a \@xhline
}

\definecolor{cvprblue}{rgb}{0.21,0.49,0.74}
\usepackage{balance}
\usepackage[pagebackref,breaklinks,colorlinks,linkcolor=cvprblue,citecolor=cvprblue,urlcolor=black]{hyperref}

\usepackage[capitalize]{cleveref}
\crefname{section}{\S}{\S\S}
\Crefname{section}{\S}{\S\S}
\crefformat{section}{\S#2#1#3}
\Crefformat{section}{\S#2#1#3}
\crefformat{subsection}{\S#2#1#3}
\Crefformat{subsection}{\S#2#1#3}
\crefformat{subsubsection}{\S#2#1#3}
\Crefformat{subsubsection}{\S#2#1#3}
\Crefname{table}{Table}{Tables}
\crefname{table}{Table}{Tables}

\def\metricsDIS{
    &$F_{\beta}^{x}\uparrow$
    &$F_{\beta}^{\omega}\uparrow$
    &$\mathcal{M}\downarrow$
    &$S_m\uparrow$
    &$E_{\phi}^{m}\uparrow$
    &$HCE_{\gamma}\downarrow$
}
\def\metricsCOD{
    &$S_m\uparrow$
    &$F_{\beta}^{\omega}\uparrow$
    &$F_{\beta}^{m}\uparrow$
    &$E_{\phi}^{m}\uparrow$
    &$E_{\phi}^{x}\uparrow$
    &$\mathcal{M}\downarrow$
}
\def\metricsSODHR{
    &$S_m\uparrow$
    &$F_{\beta}^{x}\uparrow$
    &$E_{\phi}^{m}\uparrow$
    &$\mathcal{M}\downarrow$
}
\def\metricsSODLR{
    &$S_m\uparrow$
    &$F_{\beta}^{x}\uparrow$
    &$E_{\phi}^{m}\uparrow$
    &$\mathcal{M}\downarrow$
}